\documentclass[10pt,twocolumn,letterpaper]{article}

\usepackage[OT1]{fontenc}

\usepackage[pagenumbers]{cvpr}      %

\usepackage{cuted}
\usepackage{caption}
\usepackage{algorithm}
\usepackage{algpseudocode}
\usepackage{multirow}

\definecolor{cvprblue}{rgb}{0.21,0.49,0.74}
\usepackage[pagebackref,breaklinks,colorlinks,allcolors=cvprblue]{hyperref}

\title{SparseGraph-I2P: Consistency-Guided Hypothesis Selection for Efficient Image-to-Point-Cloud Registration}

\author{Ziyang Yu\textsuperscript{1},
Xiang Li\textsuperscript{2},
Qiong Chang\textsuperscript{1},
Jun Miyazaki\textsuperscript{1}\\
\textsuperscript{1}School of Computing, Institute of Science Tokyo\\
\textsuperscript{2}School of Electronic Science \& Engineering, Nanjing University\\
{\ttfamily\footnotesize yu.ziyang.2001@gmail.com;652022230010@smail.nju.edu.cn;q.chang@c.titech.ac.jp;miyazaki@comp.isct.ac.jp}
}

\begin{document}
\raggedbottom
\setcounter{dbltopnumber}{3}
\maketitle

\begin{strip}
    \centering
    \includegraphics[width=\textwidth]{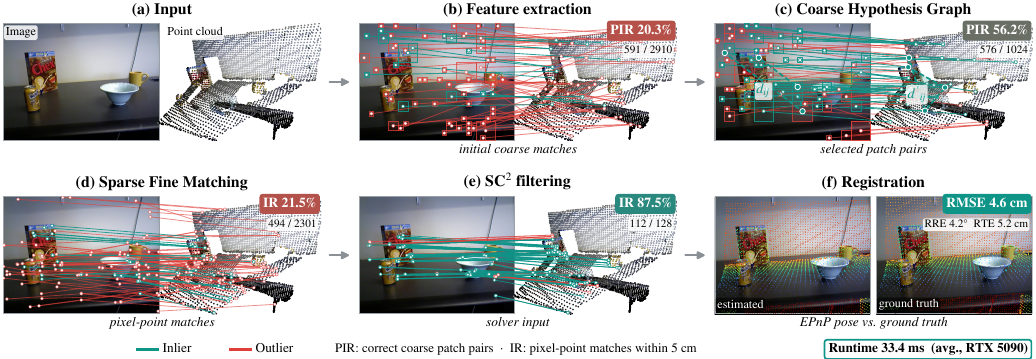}
    \captionsetup{hypcap=false}
    \captionof{figure}{SparseGraph-I2P on an RGB-D Scenes V2 test pair. The Coarse Hypothesis Graph raises
    the share of correct coarse patch pairs (PIR) from 20.3\% to 56.2\%, and SC$^2$ filtering passes
    112 inliers among 128 correspondences to a single EPnP solve (4.6\,cm RMSE).}
    \captionsetup{hypcap=true}
    \label{fig:teaser}
    \vspace{-1.2em}
\end{strip}

\begin{abstract}
Coarse-to-fine image-to-point-cloud (I2P) registration risks discarding
correct but ambiguous patch pairs, while retaining more candidates repeats
fine matching over overlapping patches. SparseGraph-I2P aims to improve
registration recall by increasing the inlier ratio at the pose solver,
using sensor or monocular depth. A Coarse Hypothesis
Graph combines appearance with depth-derived first- and second-order
compatibility to select a fixed budget of patch pairs. Sparse Fine Matching
decodes each queried pixel once and skips pixel--point comparisons covered
by an ancestor patch paired with the same point-cloud region. With sensor
depth, depth-aware filtering ranks correspondences by SC$^2$ spectral
support for a single EPnP solve. With monocular depth, depth or inverse depth
is calibrated from correspondence geometry before filtering.
SparseGraph-I2P achieves the highest registration recall and lowest latency
among the compared methods on RGB-D Scenes V2 and 7-Scenes with sensor
depth. With predicted depth, it retains the highest recall on RGB-D Scenes
V2 at an additional calibration cost. Code will be released.
\end{abstract}

\section{Introduction}
\label{sec:intro}

Image-to-point-cloud (I2P) registration estimates the camera pose of an image relative to a 3D point cloud, enabling camera relocalization and image–map alignment. In these applications, depth-equipped robots and AR devices can acquire depth alongside the query image, providing geometric observations that can help establish correspondences with the 3D map. Coarse-to-fine methods first match image and point-cloud patches, then expand selected patch pairs into pixel–point correspondences for pose estimation \cite{li2023matr}. However, cross-modal differences, viewpoint changes, and partial overlap can make correct and incorrect patch pairs difficult to distinguish by descriptor scores alone. Coarse selection must therefore balance retaining potentially correct matches against the cost of expanding unreliable candidates.

Selecting too few candidates can discard correct patch pairs before their
fine correspondences are examined; retaining more candidates preserves
alternatives but can also increase outlier contamination and matching cost.
A lower inlier ratio makes sampling an all-inlier subset less likely under
a fixed RANSAC budget. Geometric
filtering~\cite{bai2021pointdsc,chen2022sc2pcr} can reject outliers after
matching, but cannot recover inliers lost during coarse selection or avoid
the computation already spent on rejected matches. Keeping more coarse
candidates therefore does not by itself resolve the tradeoff between
inlier retention, outlier contamination, and computation. In 2D3D-MATR,
dense decoding computes descriptors even for unqueried pixels, while nested
patches paired with the same point-cloud region repeat pixel--point
comparisons. These costs limit the efficiency of exploring alternative
matches. These limitations motivate using the available depth to assess
geometric consistency before coarse candidates are discarded, together with
efficient expansion of the selected pairs and fine-level outlier rejection
before pose estimation.

We propose \textbf{SparseGraph-I2P}, which builds on a 2D3D-MATR backbone
to improve registration recall (RR) by delivering a higher inlier ratio
(IR) to the pose solver (Figure~\ref{fig:teaser}). Geometric consistency
guides coarse selection to retain promising patch pairs, then filters their
fine correspondences using SC$^2$ spectral support~\cite{chen2022sc2pcr}.
Sparse execution makes this process efficient by sharing pixel descriptors
and comparison blocks across nested patches. Higher IR reduces outlier
contamination, but reliable registration also requires sufficient inlier
coverage. Sensor depth permits a single
EPnP solve~\cite{lepetit2009epnp}; predicted depth requires correspondence-driven
calibration and robust pose estimation. On RGB-D Scenes V2 and 7-Scenes,
high solver-input IR accompanies the highest RR as well as the lowest
latency among six compared methods, all given sensor depth
(four use it only as a validity mask). With DAv2/DA3 predictions, our method
also achieves the highest RR on RGB-D Scenes V2.

The main contributions are as follows:
\begin{itemize}[leftmargin=*,nosep]
    \item SparseGraph-I2P uses a Coarse Hypothesis Graph to combine appearance and depth-derived compatibility for coarse selection within a fixed budget.
    \item Sparse Fine Matching decodes each queried pixel once and skips ancestor-covered comparisons, preserving the pixel--point comparison domain.
    \item Depth-aware filtering combines SC$^2$ ranking with EPnP for sensor depth, or correspondence-driven calibration and USAC-MSAC for predicted depth.
    \item Among compared methods, we achieve the highest RR and lowest latency on both indoor benchmarks with sensor depth, and the highest RGB-D Scenes V2 RR with predicted depth.
\end{itemize}

\section{Related Work}
\label{sec:related}

\subsection{Image-to-Point-Cloud Matching}
Early methods learn cross-modal descriptors~\cite{feng2019matchnet,wang2021p2net}
or model visibility and overlap~\cite{li2021deepi2p,ren2023corri2p}.
2D3D-MATR~\cite{li2023matr} and CoFiI2P~\cite{kang2024cofii2p} use
coarse-to-fine matching, while VP2P-Match~\cite{zhou2023vp2p}
and MinCD-PnP~\cite{an2025mincdpnp} improve geometric supervision.
FreeReg~\cite{wang2024freereg} and Diff$^2$I2P~\cite{mu2025diff2i2p} exploit
pretrained diffusion priors; GLASS~\cite{cheng2026glass} incorporates
surface-normal cues and graph-distribution consistency.
These methods primarily improve cross-modal representations and
correspondence quality. Relocalization and image--map alignment on robots
and AR devices, however, must also keep pace with the camera, so latency
matters alongside accuracy. Contextual image matchers~\cite{sarlin2020superglue,sun2021loftr} have been accelerated through adaptive computation and efficient
attention~\cite{lindenberger2023lightglue,wang2024efficientloftr}, and computation reuse and sparse execution reduce other vision
workloads~\cite{ragankelley2013halide,tang2022torchsparse,feng2020mesorasi}. This work instead lets coarse hypotheses allocate fine matching and removes two I2P-specific redundancies: decoding unqueried pixels and repeating nested pixel--point comparisons.

\begin{figure*}[!t]
    \centering
    \includegraphics[width=\textwidth]{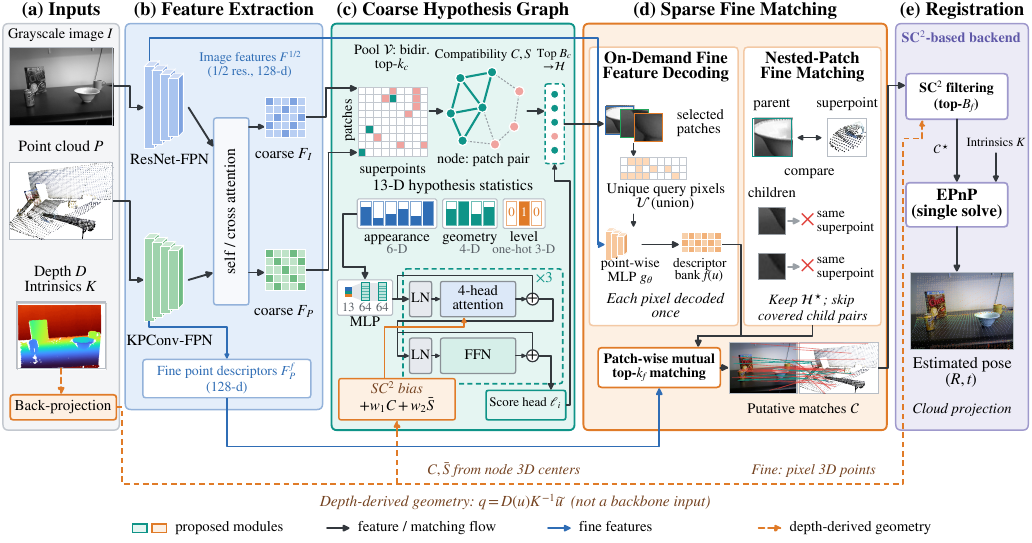}
    \caption{Overview of SparseGraph-I2P with sensor depth. (a)~Image $I$, point
    cloud $P$, depth $D$, and camera intrinsics $K$. (b)~A frozen 2D3D-MATR
    backbone provides coarse descriptors $F_I,F_P$ and fine features
    $F^{1/2},F_P^f$. (c)~The Coarse Hypothesis Graph scores the bidirectional
    top-$k_c$ pool $\mathcal{V}$ from appearance statistics and depth-derived
    compatibility $C,S$ and keeps the top $B_c$ pairs $\mathcal{H}$. (d)~Sparse
    Fine Matching decodes the queried pixels $\mathcal{U}$ once, skips covered
    child pairs ($\mathcal{H}^\star$), and outputs putative matches
    $\mathcal{C}$. (e)~SC$^2$ filtering keeps the top $B_f$ correspondences
    $\mathcal{C}^\star$ for a single EPnP solve with $K$. Depth (dashed) enters
    through validity masks and back-projection (Eq.~\eqref{eq:backprojection}); predicted
    depth uses the calibration and robust solver of
    Sec.~\ref{sec:predicted-backend}.}
    \label{fig:pipeline}
\end{figure*}

\subsection{Geometric Consistency for Correspondence Selection}
Point-cloud registration uses sampling~\cite{fischler1981ransac,chum2005prosac},
certifiable optimization~\cite{yang2021teaser}, learned
confidence~\cite{choy2020dgr,bai2021pointdsc}, and local pose
hypotheses~\cite{qin2022geotransformer} to reject outliers. Compatibility
methods rank correspondences spectrally~\cite{leordeanu2005spectral}, seek
consistent subsets~\cite{mangelson2018pcm,lusk2021clipper}, or exploit
second-order support~\cite{chen2022sc2pcr} and clique
structure~\cite{zhang2023mac,zhang2024fastmac,yan2025turboreg}. Applied after
fine matching, these methods cannot recover candidates discarded earlier.

For I2P, GraphI2P~\cite{bie2025graphi2p} selects virtual-to-LiDAR
correspondences, Top-I2P~\cite{an2025topi2p} reasons about topology, and
Hg-I2P~\cite{an2026hgi2p} uses heterogeneous region graphs for feature
adaptation and pruning. Ex-Sim(3)-Reg~\cite{an2026exsim3} handles monocular
depth noise through compatibility graphs over scale hypotheses.
SparseGraph-I2P uses compatibility-biased attention, as in PointDSC, to
score coarse patch-pair hypotheses before fine matching, then reuses
consistency for fine filtering. Since rigid-distance compatibility requires
metric geometry, predicted depth is scale-corrected before coarse selection
and calibrated in depth or inverse depth before fine filtering.

\section{Method}
\label{sec:method}

\subsection{Problem Formulation and Overview}
Given an image $I$, point cloud $P$, camera intrinsics $K$, and aligned
image-side depth $D$, the task is to estimate a rigid pose $(R,t)$ mapping
point-cloud coordinates into the camera frame. Given a selected set of
pixel--point correspondences $\mathcal{C}^{\star}=\{(u_i,p_i)\}$, the
registration objective is to minimize reprojection error:
\begin{equation}
    (R,t)=\underset{R,t}{\arg\min}
    \sum_{(u_i,p_i)\in\mathcal{C}^{\star}}
    \left\|u_i-\pi\!\left(K(Rp_i+t)\right)\right\|_2^2,
    \label{eq:registration}
\end{equation}
where $R\in\mathrm{SO}(3)$ and $t\in\mathbb{R}^{3}$ are the rotation and
translation, and $\pi([x,y,z]^\top)=[x/z,y/z]^\top$ converts homogeneous
coordinates to pixel coordinates. Depth in meters lifts a pixel to
\begin{equation}
    q_i = D(u_i)K^{-1}\widetilde{u}_i,
    \label{eq:backprojection}
\end{equation}
where $\widetilde{u}_i$ is its homogeneous coordinate. The pair $(q_i,p_i)$
denotes the image-side and cloud-side 3D representatives of a correspondence
hypothesis, with $i,j$ indexing hypotheses at either resolution: coarse
hypotheses use patch representatives and fine hypotheses individual points.

SparseGraph-I2P uses coarse selection and fine filtering to improve the
correspondences reaching the pose solver (Figure~\ref{fig:pipeline}). A frozen
2D3D-MATR backbone~\cite{li2023matr} extracts cross-modal descriptors (a--b):
ResNet-FPN and KPConv-FPN encode the image and point cloud, and self/cross
attention produces coarse descriptors $F_I$ and $F_P$. The image branch also
provides a half-resolution feature map $F^{1/2}$, and the point branch fine
point descriptors $F_P^f$. The Coarse Hypothesis Graph (Sec.~\ref{sec:coarse})
selects up to $B_c$ coarse patch pairs (c), Sparse Fine Matching
(Sec.~\ref{sec:sparse}) expands them into pixel--point correspondences (d),
and SC$^2$-based filtering keeps consistent correspondences for a single EPnP
solve with sensor depth (e, Sec.~\ref{sec:backend}). Depth supplies validity
masks and geometric compatibility. The sensor-depth pipeline is described
first; Sec.~\ref{sec:predicted-backend} adds calibration and USAC-MSAC for
monocular predictions.

\subsection{Coarse Hypothesis Graph}
\label{sec:coarse}
Selecting coarse pairs by descriptor similarity discards correct but
ambiguous matches, while keeping more candidates raises fine-matching cost.
The Coarse Hypothesis Graph (Figure~\ref{fig:pipeline}c) instead scores a broad
candidate pool jointly with depth-derived rigid compatibility and keeps a
budget of at most $B_c$ patch pairs (Figure~\ref{fig:module-graph}).

\subsubsection{Graph Construction}
\begin{figure}[t]
    \centering
    \includegraphics[width=\columnwidth]{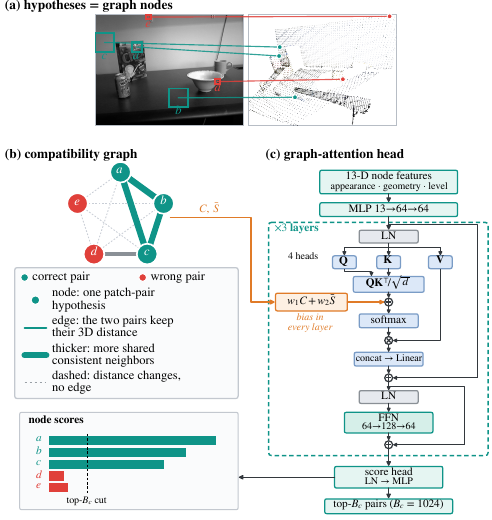}
    \caption{Coarse Hypothesis Graph on the pair of Figure~\ref{fig:teaser}.
    (a)~Each candidate patch pair is a node (green: correct; red: wrong).
    (b)~Edges join pairs whose image-side and cloud-side distances agree
    (Eq.~\eqref{eq:compatibility}); thicker edges share more compatible
    neighbors. Correct pairs form a mutually supporting cluster, while wrong
    pairs stay weakly connected. (c)~Attention biased by $C$ and $\bar{S}$ in every
    layer scores the nodes, and the top $B_c$ pairs are kept.}
    \label{fig:module-graph}
\end{figure}

Let $r$ index image patches across three pyramid levels and $s$ index
superpoints, whose point sets $\mathcal{P}_s$ partition the cloud by nearest
superpoint. The candidate set $\mathcal{V}$ is the deduplicated union of
image-to-point and point-to-image top-$k_c$ lists under the cosine similarity
of $F_I$ and $F_P$, which preserves alternatives when a correct pair ranks
highly in only one direction. Each candidate $i\in\mathcal{V}$ is a graph node
(Figure~\ref{fig:module-graph}a), with $q_i$ the mean of valid back-projected
samples of its image patch and $p_i$ its superpoint.

Rigid motion preserves distances between corresponding 3D points. Patch
representatives, however, need not denote the same physical locations under
partial visibility. Coarse distance agreement is therefore approximate
evidence, rather than an exact constraint on correct patch pairs. Following
spatial compatibility~\cite{chen2022sc2pcr}, we measure this evidence with a
tolerance:
\begin{equation}
    C_{ij}=\left[1-
    \frac{(\|q_i-q_j\|_2-\|p_i-p_j\|_2)^2}{\sigma_{ij}^2}\right]_+,
    \label{eq:compatibility}
\end{equation}
where $[x]_+=\max(x,0)$ and $C_{ii}=0$. The same definition is used at both
resolutions, with tolerance $\sigma_{ij}=\sigma_c$ for coarse hypotheses.
Thresholding gives $A_{ij}=\mathbb{1}[C_{ij}>\tau]$, and second-order support
is
\begin{equation}
    S=(AA)\odot A,
    \label{eq:secondorder}
\end{equation}
where the matrix product counts common compatible neighbors and $\odot$
retains only compatible edges. Thus $C$ measures direct distance agreement,
while $S$ measures collective support: false hypotheses may agree by chance
without sharing the compatible neighbors of a correct cluster
(Figure~\ref{fig:module-graph}b, edge widths proportional to $S_{ij}$). For
attention, $\bar{S}$ denotes $S$ normalized by its largest entry, or zero when
$S=0$.

\subsubsection{Graph-Attention Scoring}
To accommodate approximate patch geometry, the graph learns to combine
appearance and geometric support (Figure~\ref{fig:module-graph}c). Each node
contains six appearance statistics (similarity, two directional gaps, two
ranks, and mutual-top-1 membership), four geometric measures (first- and
second-order degree, spectral support, and local support), and three
pyramid-level indicators. An MLP embeds these features into 64 dimensions;
three pre-norm layers then aggregate evidence across hypotheses
with attention biased by compatibility. Each head computes
\begin{equation}
    \operatorname{softmax}\!\left(
    \frac{\mathbf{Q}\mathbf{K}^\top}{\sqrt{d}}+
    w_1 C+w_2\bar{S}\right)\mathbf{V},
    \label{eq:graph-attention}
\end{equation}
where $d$ is the head dimension, $\mathbf{Q},\mathbf{K},\mathbf{V}$ are
learned projections (bold $\mathbf{K}$ is not the intrinsics $K$), and
$w_1,w_2$ are learned for each head and layer. The feature term relates node
attributes, while $C$ and $\bar{S}$ add direct and collective support.
Geometry biases attention without masking it, so imperfect patch geometry
need not exclude useful hypotheses. After residual feed-forward blocks and
layer normalization, an MLP gives a logit $\ell_i$ per node, and up to $B_c$
highest-scoring nodes form $\mathcal{H}\subseteq\mathcal{V}$. The
candidate-pool size is thus decoupled from the fine-matching budget.

\subsection{Sparse Fine Matching}
\label{sec:sparse}
\begin{figure}[t]
    \centering
    \includegraphics[width=\columnwidth]{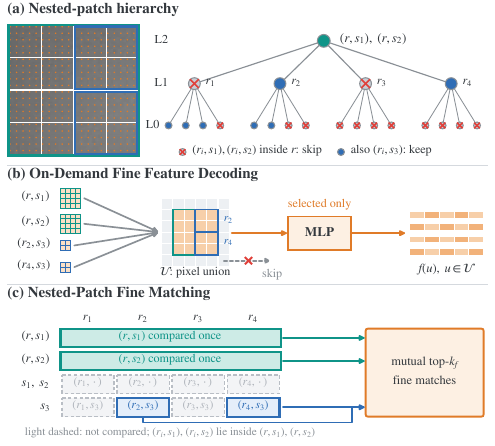}
    \caption{Sparse Fine Matching (Figure~\ref{fig:pipeline}d) on the pair of
    Figure~\ref{fig:teaser}. (a)~Image patches form a quadtree; a child pair
    is skipped when a selected ancestor covers it with the same superpoint and
    kept for a different one ($s_3$). (b)~On-Demand Fine Feature Decoding
    decodes the union of queried pixels once. (c)~Nested-Patch Fine Matching
    compares each pixel--point block once.}
    \label{fig:module-fine}
\end{figure}

Overlapping patches make fine matching redundant: descriptors are decoded
for the whole image, and nested patches paired with the same superpoint
compare the same pixel--point block again. Sparse Fine Matching
(Figure~\ref{fig:pipeline}d) organizes all image patches in one quadtree,
then decodes each queried pixel once and skips blocks already covered by a
selected ancestor (Figure~\ref{fig:module-fine}).

Each selected pair $(r,s)\in\mathcal{H}$ requests the comparison block
$U_r\times\mathcal{P}_s$, where $U_r$ is the sampled pixel set of patch $r$.
The image patches form a quadtree on a common pixel lattice
(Figure~\ref{fig:module-fine}a): level~0 is the $24\times32$ patch grid and
each coarser level halves both sides, so patch $r=(l,i,j)$ has parent
$\operatorname{par}(r)=(l{+}1,\lfloor i/2\rfloor,\lfloor j/2\rfloor)$ and $U_{\operatorname{par}(r)}$ is the
union of the pixel sets of its four children. We write $\mathcal{A}(r)$ for the
ancestors of $r$, obtained by repeating $\operatorname{par}$. The tree indexes queried pixels and covered blocks once per image.

\subsubsection{On-Demand Fine Feature Decoding}
The convolutional decoder of 2D3D-MATR produces a full descriptor map.
To enable independent pixel queries, we replace it with a point-wise MLP
trained by distillation and fine-matching supervision (Sec.~\ref{sec:loss}).
This is a learned approximation to the original decoder. We form the
query union $\mathcal{U}=\bigcup_{(r,s)\in\mathcal{H}}U_r$ on the lattice
and batch-decode its pixels once (Figure~\ref{fig:module-fine}b):
\begin{equation}
    f(u)=\operatorname{norm}\!\left(
    g_\theta(\operatorname{bilinear}(F^{1/2},u))\right),
    \quad u\in\mathcal{U},
    \label{eq:sparse}
\end{equation}
where $g_\theta$ is the point-wise decoder and $\operatorname{norm}$ denotes
$\ell_2$ normalization. All patches share a
descriptor bank indexed by global pixel IDs, so a pixel requested by a parent
and several children is decoded once. Decoding cost therefore scales with
$|\mathcal{U}|$ rather than with the image size or the number of pairs.

\subsubsection{Nested-Patch Fine Matching}
Matching every selected pair independently repeats comparisons whenever
nested patches share a superpoint. Since $U_r\subset U_a$ for every
$a\in\mathcal{A}(r)$, a pair $(r,s)$ is discarded if a selected ancestor pair
$(a,s)$ already covers its block (Figure~\ref{fig:module-fine}c):
\begin{equation}
    \mathcal{H}^{\star}=\{(r,s)\in\mathcal{H}:
    (a,s)\notin\mathcal{H}\ \ \forall a\in\mathcal{A}(r)\}.
    \label{eq:nested}
\end{equation}
The retained set preserves the comparison domain,
\begin{equation}
    \bigcup_{(r,s)\in\mathcal{H}^{\star}} U_r\times\mathcal{P}_s
    =\bigcup_{(r,s)\in\mathcal{H}} U_r\times\mathcal{P}_s,
    \label{eq:comparison-domain}
\end{equation}
so every requested pixel--point pair is evaluated once. This preserves the
comparison domain, but not necessarily the output correspondences: mutual
top-$k_f$ rankings are computed within each retained block and can change
when child blocks are removed. Comparison cost scales with
$\sum_{(r,s)\in\mathcal{H}^{\star}}|U_r||\mathcal{P}_s|$. Within each retained block, a pixel
$u$ and point $p$ are matched if they are mutual top-$k_f$ under
$z(u,p)=\tfrac12(1+f(u)^\top F_P^f(p))$ and $z(u,p)>\gamma$; repeats of the
same global IDs are merged into $\mathcal{C}=\{(u_i,p_i)\}$, with lifted
points $q_i$ from Eq.~\eqref{eq:backprojection}. Retained blocks are grouped by pyramid
level for batched GPU products; the Coarse Hypothesis Graph itself remains dense.

\subsection{Depth-Aware Filtering and Pose Estimation}
\label{sec:backend}
Fine correspondences still contain outliers, even inside correct coarse
pairs. The compatibility of Sec.~\ref{sec:coarse} is reused at pixel level to
filter them before pose estimation. Sensor depth uses a single EPnP solve;
predicted depth uses correspondence-driven calibration and USAC-MSAC.

\subsubsection{Sensor Depth}
Each fine correspondence in $\mathcal{C}$ is lifted by
Eq.~\eqref{eq:backprojection}, and
Eqs.~\eqref{eq:compatibility}--\eqref{eq:secondorder} give $C$, $A$, and $S$
over $\mathcal{C}$ with $\sigma_{ij}=\sigma_f$ (Figure~\ref{fig:pipeline}e).
We use a simplified SC$^2$~\cite{chen2022sc2pcr}: instead of selecting seeds
on $C$ and expanding a consensus set per seed, spectral
matching~\cite{leordeanu2005spectral} ranks all correspondences once on $S$.
For $|\mathcal{C}|\ge4$, starting from
$v^{(0)}=\mathbf{1}/\sqrt{|\mathcal{C}|}$, $T=10$ power iterations
\begin{equation}
    v^{(n+1)}=\frac{S v^{(n)}}{\|S v^{(n)}\|_2}
    \label{eq:fine-spectral}
\end{equation}
estimate the leading eigenvector of $S$. If the normalization denominator
is below $10^{-12}$, we use uniform scores. The top
$\min(B_f,|\mathcal{C}|)$ correspondences form
\begin{equation}
    \mathcal{C}^{\star}=\{(u_i,p_i)\in\mathcal{C}:
    i\in\operatorname{TopK}_{\min(B_f,|\mathcal{C}|)}(v^{(T)})\}.
    \label{eq:fine-selection}
\end{equation}
Depth is used to select correspondences; the final pose is estimated from
their original 2D pixels and 3D cloud points. We apply
EPnP~\cite{lepetit2009epnp} once to $\mathcal{C}^{\star}$, without RANSAC,
using its algebraic fit to approximate Eq.~\eqref{eq:registration}. This
choice relies on the quality and geometric coverage of the selected
correspondences. Fewer than four correspondences or an unsuccessful EPnP
solve returns no pose.

\subsubsection{Predicted Depth}
\label{sec:predicted-backend}
Monocular predictions distort the geometry through scale and offset errors.
Predicted depth supplies the valid-pixel and valid-patch masks and the coarse
patch representatives. Before graph selection, we estimate scale from the dominant mode of
cloud-to-image distance ratios in the appearance-based candidate pool.
These pose-invariant ratios rescale the coarse representatives. After fine
matching, affine models in depth and inverse depth are initialized by
pairwise distance-agreement voting. We refine and select the model using
the leading eigenvalue of its soft second-order compatibility matrix,
favoring collectively consistent correspondences. The calibrated depth
$\widehat{D}$ replaces $D$ in Eq.~\eqref{eq:backprojection} for fine filtering.
Distance-adaptive tolerances and spatially spread one-to-one matches
accommodate residual errors, and USAC-MSAC~\cite{raguram2013usac} estimates
the pose from the selected original pixel--point pairs. Neither sensor depth
nor ground-truth poses enter this inference path.

\subsection{Loss Function}
\label{sec:loss}
We freeze the 2D3D-MATR backbone and train the graph and decoder separately.
The graph uses class-weighted binary cross-entropy over the candidates:
\begin{equation}
    \mathcal{L}_{\mathrm{graph}}=-\frac{1}{|\mathcal{V}|}\sum_{i\in\mathcal{V}}
    \left[\alpha y_i\log\hat{y}_i+(1-y_i)\log(1-\hat{y}_i)\right],
    \label{eq:graph-loss}
\end{equation}
where $\hat{y}_i=\operatorname{sigmoid}(\ell_i)$, $y_i$ indicates nonzero
ground-truth patch overlap in both directions, and $\alpha$ is the
training-set ratio of negative to positive candidates. The decoder loss is
\begin{equation}
    \mathcal{L}_{\mathrm{dec}}=\mathcal{L}_{\mathrm{fine}}
    +\lambda_d\mathcal{L}_{\mathrm{distill}},\qquad \lambda_d=1,
    \label{eq:decoder-loss}
\end{equation}
where 2D3D-MATR's fine circle loss $\mathcal{L}_{\mathrm{fine}}$~\cite{li2023matr}
supervises pixel--point discrimination using ground-truth 2D/3D distances,
and $\mathcal{L}_{\mathrm{distill}}$ transfers the frozen dense decoder's
descriptors $f_T$ by averaging $1-\cos(f(u),f_T(u))$ over sampled training
pixels.

\section{Experiments}
\label{sec:experiments}

All evaluations run on an otherwise idle machine with a 32\,GB NVIDIA RTX 5090
GPU and an AMD Ryzen 7 9850X3D CPU.
We follow the 2D3D-MATR benchmark~\cite{li2023matr}. RGB-D Scenes V2 uses
scenes 1--9 for training, scene 10 for validation, and scenes 11--14 (497
pairs) for testing; 7-Scenes uses the standard split with 2304 test pairs.
Inliers have 3D residual below 5\,cm. IR is measured on the solver input,
FMR counts pairs whose candidate IR before fine filtering is at least 10\%,
and RR counts pairs with registration RMSE below 10\,cm. Percentages are
averaged equally across scenes, and bold marks the best values, including
ties. We use $k_c=3$, $B_c=1024$,
$\sigma_c=0.1$\,m, $\tau=0.1$, and $\gamma=0.75$, with
$(k_f,\sigma_f,B_f)=(2,6.25\,\mathrm{mm},128)$ on RGB-D Scenes V2 and
$(1,12.5\,\mathrm{mm},64)$ on 7-Scenes. Our results average three
graph-head seeds (RR std 0.7/0.2 points); latency and ablations use one seed.

\begin{figure*}[!t]
    \centering
    \begin{minipage}[t]{\dimexpr(\textwidth-\columnsep)/2\relax}
        \centering\footnotesize
\captionof{table}{Registration on RGB-D Scenes V2 (\%).}
\label{tab:rgbd-evaluation}
\setlength{\tabcolsep}{1pt}
\renewcommand{\arraystretch}{1.0}
\begin{tabular*}{\columnwidth}{@{\extracolsep{\fill}}lrrrrr@{}}
\toprule
Method & Scene 11 & Scene 12 & Scene 13 & Scene 14 & Mean \\
\midrule
\multicolumn{6}{c}{Inlier Ratio $\uparrow$} \\
\midrule
2D3D-MATR & 32.8 & 34.4 & 39.3 & 23.3 & 32.4 \\
FreeReg & 43.1 & 46.5 & 53.4 & 33.6 & 44.2 \\
Diff$^2$I2P & 52.7 & 58.4 & 17.8 & 8.5 & 34.3 \\
GLASS & 51.5 & 52.2 & 35.2 & 30.2 & 42.3 \\
Hg-I2P & 33.9 & 38.9 & 43.1 & 28.9 & 36.2 \\
Ex-Sim(3) & 64.9 & 63.4 & 73.4 & 51.7 & 63.4 \\
\cmidrule(lr){1-6}
\textbf{Ours} & \textbf{95.4} & \textbf{94.6} & \textbf{90.5} & \textbf{75.4} & \textbf{89.0} \\
\midrule
\multicolumn{6}{c}{Feature Matching Recall $\uparrow$} \\
\midrule
2D3D-MATR & 98.6 & 98.0 & 88.7 & 79.2 & 91.1 \\
FreeReg & 98.6 & \textbf{100.0} & \textbf{100.0} & \textbf{88.9} & \textbf{96.9} \\
Diff$^2$I2P & \textbf{100.0} & \textbf{100.0} & 75.3 & 32.7 & 77.0 \\
GLASS & \textbf{100.0} & \textbf{100.0} & 89.7 & 82.7 & 93.1 \\
Hg-I2P & \textbf{100.0} & \textbf{100.0} & 92.8 & 85.0 & 94.4 \\
Ex-Sim(3) & 98.6 & 95.1 & 90.7 & 78.8 & 90.8 \\
\cmidrule(lr){1-6}
\textbf{Ours} & 97.2 & 96.7 & 86.6 & 54.7 & 83.8 \\
\midrule
\multicolumn{6}{c}{Registration Recall $\uparrow$} \\
\midrule
2D3D-MATR & 61.1 & 56.9 & 53.6 & 50.0 & 55.4 \\
FreeReg & 83.3 & \textbf{89.2} & 86.6 & 68.6 & 81.9 \\
Diff$^2$I2P & 86.1 & 78.4 & 9.3 & 14.2 & 47.0 \\
GLASS & 86.1 & 84.3 & 70.1 & 70.4 & 77.7 \\
Hg-I2P & 63.9 & 67.6 & 51.5 & 55.8 & 59.7 \\
Ex-Sim(3) & 77.8 & 63.7 & 63.9 & 46.0 & 62.9 \\
\cmidrule(lr){1-6}
\textbf{Ours} & \textbf{90.7} & 88.6 & \textbf{91.8} & \textbf{71.5} & \textbf{85.6} \\
\bottomrule
\end{tabular*}
    \end{minipage}\hfill
    \begin{minipage}[t]{\dimexpr(\textwidth-\columnsep)/2\relax}
        \centering\footnotesize
\captionof{table}{Registration on 7-Scenes (\%).}
\label{tab:7scenes-evaluation}
\setlength{\tabcolsep}{1pt}
\renewcommand{\arraystretch}{1.0}
\begin{tabular*}{\columnwidth}{@{\extracolsep{\fill}}lrrrrrrrr@{}}
\toprule
Method & Chess & Fire & Heads & Office & Pump. & Kitch. & Stairs & Mean \\
\midrule
\multicolumn{9}{c}{Inlier Ratio $\uparrow$} \\
\midrule
2D3D-MATR & 72.1 & 66.1 & 31.3 & 60.7 & 50.2 & 52.5 & 18.1 & 50.1 \\
FreeReg & 52.9 & 50.9 & 50.8 & 46.6 & 28.2 & 41.4 & 34.5 & 43.6 \\
Diff$^2$I2P & 74.1 & 68.0 & 52.8 & 67.8 & 50.1 & 56.9 & 20.9 & 55.8 \\
GLASS & 77.6 & 69.1 & 44.1 & 67.1 & 53.5 & 56.3 & 19.5 & 55.3 \\
Hg-I2P & 73.4 & 68.0 & 43.6 & 65.4 & 52.3 & 54.5 & 18.6 & 53.7 \\
Ex-Sim(3) & 84.9 & 81.6 & 67.0 & 73.9 & 59.6 & 61.0 & 27.1 & 65.0 \\
\cmidrule(lr){1-9}
\textbf{Ours} & \textbf{100.0} & \textbf{99.3} & \textbf{89.7} & \textbf{99.3} & \textbf{82.9} & \textbf{88.7} & \textbf{37.0} & \textbf{85.3} \\
\midrule
\multicolumn{9}{c}{Feature Matching Recall $\uparrow$} \\
\midrule
2D3D-MATR & \textbf{100.0} & 99.6 & 98.6 & \textbf{100.0} & 92.4 & 95.7 & 58.1 & 92.1 \\
FreeReg & 91.6 & 95.8 & 95.9 & 96.0 & 83.3 & 96.0 & \textbf{93.2} & 93.1 \\
Diff$^2$I2P & \textbf{100.0} & 99.3 & 98.6 & 99.8 & 92.4 & \textbf{99.0} & 63.5 & 93.2 \\
GLASS & \textbf{100.0} & \textbf{100.0} & 98.6 & \textbf{100.0} & 90.3 & 95.7 & 75.7 & \textbf{94.3} \\
Hg-I2P & \textbf{100.0} & \textbf{100.0} & \textbf{100.0} & \textbf{100.0} & \textbf{94.8} & 96.3 & 66.2 & 93.9 \\
Ex-Sim(3) & \textbf{100.0} & 99.6 & 97.3 & \textbf{100.0} & 92.4 & 95.9 & 40.5 & 89.4 \\
\cmidrule(lr){1-9}
\textbf{Ours} & \textbf{100.0} & 99.7 & 97.3 & \textbf{100.0} & 92.2 & 96.6 & 57.7 & 91.9 \\
\midrule
\multicolumn{9}{c}{Registration Recall $\uparrow$} \\
\midrule
2D3D-MATR & 96.5 & 92.3 & 45.2 & 94.0 & 79.9 & 85.2 & 25.7 & 74.1 \\
FreeReg & 85.0 & 83.4 & 84.9 & 88.2 & 61.1 & 88.6 & \textbf{58.1} & 78.5 \\
Diff$^2$I2P & 99.7 & 93.4 & 86.3 & 98.0 & 81.6 & 94.1 & 33.8 & 83.8 \\
GLASS & 99.0 & 91.2 & 80.8 & 94.2 & 78.8 & 87.2 & 27.0 & 79.7 \\
Hg-I2P & 98.6 & 97.1 & 75.3 & 98.0 & 81.9 & 90.2 & 36.5 & 82.5 \\
Ex-Sim(3) & 97.2 & 94.0 & 82.2 & 95.5 & 80.6 & 85.0 & 23.0 & 79.6 \\
\cmidrule(lr){1-9}
\textbf{Ours} & \textbf{100.0} & \textbf{98.2} & \textbf{98.6} & \textbf{98.6} & \textbf{89.5} & \textbf{95.5} & 37.8 & \textbf{88.3} \\
\bottomrule
\end{tabular*}
    \end{minipage}
    \par\smallskip
    \includegraphics[width=\textwidth]{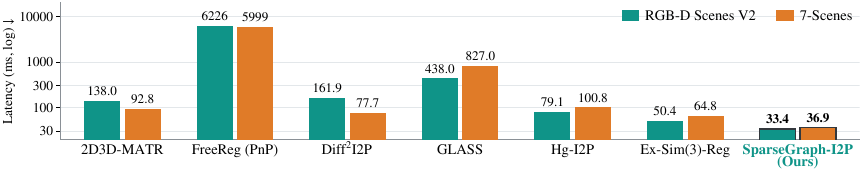}
    \captionsetup{skip=3pt}
    \caption{Registration latency on RGB-D Scenes V2 and 7-Scenes (lower
    is better).}
    \label{fig:benchmark-latency}
\end{figure*}

All methods in Tables~\ref{tab:rgbd-evaluation} and~\ref{tab:7scenes-evaluation}
use sensor depth with their own registration backends: 2D3D-MATR,
Diff$^2$I2P, GLASS, and Hg-I2P as a validity mask, and FreeReg and
Ex-Sim(3)-Reg~\cite{an2026exsim3} in place of ZoeDepth and DAv2, which
improves both (Figure~\ref{fig:predicted-depth}a). 
Latency (Figure~\ref{fig:benchmark-latency}) covers forward, filtering, and
pose estimation.

\subsection{Comparison with State-of-the-Art Methods}
\label{sec:benchmark-comparison}
\label{sec:rgbd-evaluation}
\label{sec:7scenes-evaluation}

On RGB-D Scenes V2 (Table~\ref{tab:rgbd-evaluation}), SparseGraph-I2P
attains the highest mean RR (85.6\%), 3.7 points above FreeReg (81.9\%),
and the highest RR in three of the four scenes. Its solver-input
IR (89.0\%, 128 kept by SC$^2$) exceeds baseline IR on full candidate sets
(at most 63.4\%, Ex-Sim(3)-Reg), and
it runs at 33.4\,ms per pair versus 50.4\,ms for the fastest baseline
(Figure~\ref{fig:benchmark-latency}). Its FMR (83.8\%), measured before SC$^2$
filtering, is below most baselines'; on Scene 14 the filter registers pairs
with candidate IR below 10\%, so RR (71.5\%) exceeds FMR (54.7\%).

On 7-Scenes (Table~\ref{tab:7scenes-evaluation}), it reaches 88.3\% RR,
4.5 points above Diff$^2$I2P (83.8\%), with solver-input IR 85.3\% (64 kept) versus
at most 65.0\% and latency 36.9\,ms versus 64.8\,ms for the fastest
baseline. Stairs is the exception: FreeReg reaches 58.1\% RR versus 37.8\%;
all others stay below 40\%.
Higher solver-input IR accompanies higher mean RR on both datasets,
consistent with our design goal. These IR values reflect each method's
filtering and set size, rather than raw matching quality alone.

\begin{figure*}[!t]
    \centering
    \includegraphics[width=\textwidth]{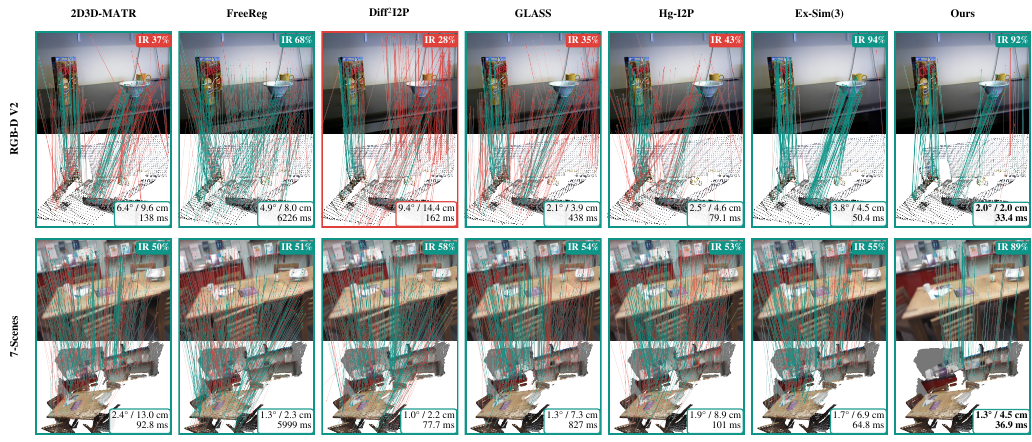}
    \caption{Solver-input correspondences of every method on one test pair per
    dataset (green: inlier within 5\,cm; red: outlier; up to 250 drawn per
    baseline, all 128/64 for SparseGraph-I2P). Badges give solver-input IR; the box at the
    lower right gives RRE\,/\,RTE of the estimated pose (green frame: RMSE
    $<10$\,cm) and the method's mean latency on the dataset
    (Figure~\ref{fig:benchmark-latency}).
    All methods use sensor depth.}
    \label{fig:qualitative}
\end{figure*}

\begin{figure*}[t]
    \centering
    \includegraphics[width=\textwidth]{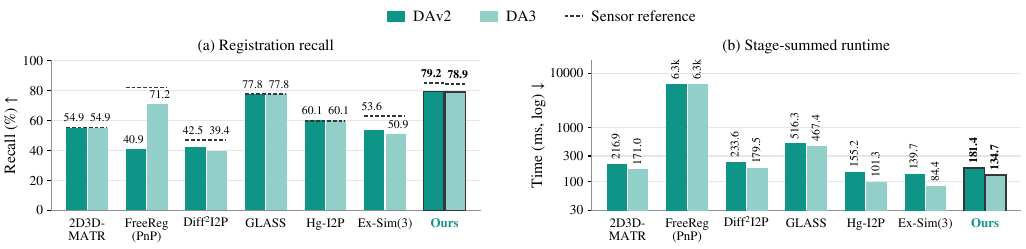}
    \captionsetup{skip=3pt}
    \caption{Predicted-depth registration on RGB-D Scenes V2: (a) scene-mean recall;
    (b) sum of measured depth-prediction, forward, filtering and solving
    times (log scale; disk IO and offline MobileSAM excluded). Solid/light
    bars use DAv2/DA3 throughout inference. Dashed marks in (a) show
    sensor-depth references. For ours, these marks retain prediction-based
    coarse hypotheses and fine candidates, use sensor depth and its fine-filter settings,
    and keep USAC-MSAC. Stage sums are not directly timed end-to-end latency.}
    \label{fig:predicted-depth}
\end{figure*}

In Figure~\ref{fig:qualitative}, baseline solver-input IR ranges from 28\%
to 94\% and PnP-RANSAC registers 11 of the 12 baseline cells, whereas
SparseGraph-I2P passes 92\% and 89\% inliers to a single EPnP solve. 

\subsection{Ablation Studies}
\label{sec:ablations}
In Table~\ref{tab:ablation-components}, (a)--(b) use the original
convolutional decoder and (c)--(g) the distilled point-wise MLP.
Replacing PnP-RANSAC with SC$^2$/EPnP, (a)$\to$(b), raises RGB-D/7-Scenes RR
from 55.4/74.1\% to 73.7/86.5\% and cuts latency from 138.0/92.8\,ms to
30.0/37.0\,ms. Variants (c) and (g) differ only in coarse selection: the
Coarse Hypothesis Graph raises RR from 69.4/84.4\% (hand-crafted SC$^2$) to
84.8/88.2\%, while solver-input IR rises from 78.4/81.8\% to 88.8/85.3\%.
Over (b), the full system adds 11.1/1.7 RR points.

\begin{table*}[!t]
\centering\footnotesize
\caption{Sensor-depth ablation of the three components (\%, ms; graph-head seed 7351). Without the learned graph,
(a)--(b) use 2D3D-MATR coarse matching; (a) uses PnP-RANSAC.}
\label{tab:ablation-components}
\setlength{\tabcolsep}{2pt}
\renewcommand{\arraystretch}{1.15}
\begin{tabular*}{\textwidth}{@{\extracolsep{\fill}}lccccrrrrrrrr@{}}
\toprule
 & \multirow{2}{*}{Coarse Hypothesis Graph} & \multicolumn{2}{c}{Sparse Fine Matching} & \multirow{2}{*}{Depth-aware Filtering} & \multicolumn{4}{c}{RGB-D V2} & \multicolumn{4}{c}{7-Scenes} \\
\cmidrule(lr){3-4}\cmidrule(lr){6-9}\cmidrule(lr){10-13}
Variant & & On-demand & Nested & & IR$\uparrow$ & FMR$\uparrow$ & RR$\uparrow$ & Lat.$\downarrow$ & IR$\uparrow$ & FMR$\uparrow$ & RR$\uparrow$ & Lat.$\downarrow$ \\
\midrule
(a) & -- & -- & -- & -- & 32.4 & \textbf{91.1} & 55.4 & 138.0 & 50.1 & \textbf{92.1} & 74.1 & 92.8 \\
(b) & -- & -- & -- & $\checkmark$ & 81.2 & \textbf{91.1} & 73.7 & 30.0 & 83.3 & \textbf{92.1} & 86.5 & 37.0 \\
(c) & -- & $\checkmark$ & $\checkmark$ & $\checkmark$ & 78.4 & 90.5 & 69.4 & \textbf{24.8} & 81.8 & 91.1 & 84.4 & \textbf{31.9} \\
(d) & $\checkmark$ & -- & -- & $\checkmark$ & \textbf{90.3} & 85.1 & \textbf{85.0} & 46.4 & \textbf{86.1} & 91.5 & \textbf{88.4} & 42.4 \\
(e) & $\checkmark$ & -- & $\checkmark$ & $\checkmark$ & 88.8 & 83.6 & 84.8 & 35.0 & 85.3 & 92.0 & 88.2 & 38.5 \\
(f) & $\checkmark$ & $\checkmark$ & -- & $\checkmark$ & \textbf{90.3} & 85.1 & \textbf{85.0} & 43.2 & \textbf{86.1} & 91.5 & \textbf{88.4} & 41.2 \\
\midrule
(g) \textbf{Ours} & $\checkmark$ & $\checkmark$ & $\checkmark$ & $\checkmark$ & 88.8 & 83.6 & 84.8 & 33.4 & 85.3 & 92.0 & 88.2 & 36.9 \\
\bottomrule
\end{tabular*}
\end{table*}

On-Demand Fine Feature Decoding saves 3.2/1.2\,ms, (d)$\to$(f), and
1.6/1.6\,ms, (e)$\to$(g), without changing accuracy. Nested-Patch Fine
Matching saves 11.4/3.9 and 9.8/4.3\,ms, (d)$\to$(e) and (f)$\to$(g), at a
0.2-point RR cost. Together they cut latency by 28\%/13\%, to
33.4/36.9\,ms. Thus coarse selection and filtering improve IR and RR,
while sparse fine matching lowers computation cost.

\paragraph{Coarse matching budget.}
Table~\ref{tab:ablation-budget} varies $B_c$ from 256 to 2048 while keeping
graph selection, on-demand decoding, nested-patch matching, and the fine
backend fixed. Increasing $B_c$ from 1024 to 2048 raises RGB-D RR from
84.8\% to 86.1\%, but increases latency from 33.4 to 46.1\,ms; 7-Scenes RR
falls from 88.2\% to 87.6\%. We retain the validation-selected budget of
1024. This budget limits coarse patch pairs, not fine correspondences or
solver inputs.

\begin{table}[!tbp]
\centering\footnotesize
\caption{Coarse matching budget ablation (\%, ms).}
\label{tab:ablation-budget}
\setlength{\tabcolsep}{1pt}
\renewcommand{\arraystretch}{1.05}
\begin{tabular*}{\columnwidth}{@{\extracolsep{\fill}}lrrrrrrrr@{}}
\toprule
 & \multicolumn{4}{c}{RGB-D V2} & \multicolumn{4}{c}{7-Scenes} \\
\cmidrule(lr){2-5}\cmidrule(lr){6-9}
Setting & IR$\uparrow$ & FMR$\uparrow$ & RR$\uparrow$ & Lat.$\downarrow$ & IR$\uparrow$ & FMR$\uparrow$ & RR$\uparrow$ & Lat.$\downarrow$ \\
\midrule
256 & 87.9 & \textbf{92.4} & 84.3 & \textbf{28.4} & 84.4 & 91.6 & 86.9 & \textbf{35.9} \\
512 & 88.7 & 90.0 & 85.1 & 29.4 & 84.6 & 91.9 & 87.8 & 36.7 \\
768 & 88.8 & 87.1 & 85.3 & 31.8 & 85.1 & 91.9 & 88.1 & 36.6 \\
1536 & 88.9 & 75.1 & 86.0 & 38.7 & 85.0 & 91.7 & 88.1 & 37.1 \\
2048 & \textbf{89.0} & 68.1 & \textbf{86.1} & 46.1 & \textbf{85.3} & 90.8 & 87.6 & 38.7 \\
\midrule
1024 (\textbf{Ours}) & 88.8 & 83.6 & 84.8 & 33.4 & \textbf{85.3} & \textbf{92.0} & \textbf{88.2} & 36.9 \\
\bottomrule
\end{tabular*}
\end{table}

\subsection{Registration with Predicted Depth}
\label{sec:predicted-depth}

Sensor depth is replaced by the same DAv2~\cite{yang2024dav2} or
DA3~\cite{lin2025da3} predictions for all methods, without retraining
(Figure~\ref{fig:predicted-depth}). Our runs use predictions in both coarse selection and fine
filtering, with no sensor depth at inference; our dashed diagnostic
references retain prediction-based coarse hypotheses and fine candidates,
but use sensor depth and sensor-depth settings for fine filtering, keeping
USAC-MSAC. They differ from the full system of
Tables~\ref{tab:rgbd-evaluation}--\ref{tab:7scenes-evaluation}.
On RGB-D Scenes V2, SparseGraph-I2P achieves 79.2\% RR with DAv2 and 78.9\% with DA3, the
highest under either predictor, 5.8/5.6 points below these diagnostic
references (84.9/84.5\%). On 7-Scenes it reaches 81.7\%, 1.1 points
below Hg-I2P (82.8\%). Including depth
prediction, its summed stage times are 181.4/134.7\,ms (DAv2/DA3), within the 84.4--233.6\,ms
of the other four baselines and below GLASS and FreeReg.

\section{Conclusion}
\label{sec:conclusion}
SparseGraph-I2P targets higher registration recall through higher
solver-input IR. The Coarse Hypothesis Graph retains geometrically supported
patch pairs, and depth-aware filtering suppresses fine-match outliers.
Sparse Fine Matching removes redundant decoding and comparisons.
Together, these components improve sensor-depth registration recall at low
latency; correspondence-driven calibration extends the approach to predicted
depth. Future work will investigate cheaper calibration and better inlier
coverage in ambiguous scenes.

\section*{LLM Usage Statement}
Claude, OpenAI Codex, and Qwen Image 2.1 were used as assistive tools for
polishing the manuscript's language and assisting with code development.
OpenAI Codex additionally assisted with drafting and revising the Introduction,
Method, experimental analysis, Discussion, and Conclusion, migrating the
manuscript to the \LaTeX{} template, and drafting this statement.

{\small
    \bibliographystyle{ieeenat_fullname}
    \bibliography{references}
}

\clearpage
\appendix
\maketitlesupplementary
\setcounter{equation}{0}
\setcounter{algorithm}{0}
\renewcommand{\thetable}{S\arabic{table}}
\renewcommand{\thefigure}{S\arabic{figure}}
\renewcommand{\thealgorithm}{S\arabic{algorithm}}
\renewcommand{\theequation}{S\arabic{equation}}
\renewcommand{\thesection}{\Alph{section}}
\setcounter{table}{0}
\setcounter{figure}{0}

This supplement first defines the evaluation metrics
(Section~\ref{sec:supp-metrics}), the datasets and splits
(Section~\ref{sec:supp-datasets}), and the baseline configurations
(Section~\ref{sec:supp-baselines}). Section~\ref{sec:supp-implementation}
gives the implementation details of SparseGraph-I2P, and
Section~\ref{sec:supp-experiments} reports rotation and translation errors,
and outdoor registration on KITTI-DC.
Section~\ref{sec:supp-predicted} derives and
evaluates the predicted-depth path of Section~\ref{sec:backend}, and
Section~\ref{sec:supp-gallery} shows additional qualitative results.

\section{Evaluation Metrics}
\label{sec:supp-metrics}
The indoor metrics follow the 2D3D-MATR benchmark~\cite{li2023matr} and are computed
by one evaluation script on the correspondences each method outputs. For a
test pair with point cloud $P$, ground-truth pose $(\bar R,\bar t)$ from the
cloud to the camera, and pixel--point correspondences
$\mathcal C=\{(u_i,p_i)\}$, each pixel is lifted with the sensor depth as
$q_i=D(u_i)K^{-1}\widetilde u_i$.
The outdoor thresholds and error statistics for KITTI-DC are specified
separately in Section~\ref{sec:supp-kitti-dc}.

\paragraph{Inlier Ratio (IR).}
A correspondence is an inlier if its 3D residual under the ground truth is
below $\tau_c=5$\,cm:
\begin{equation}
    \mathrm{IR}(\mathcal C)=\frac{1}{|\mathcal C|}\sum_{(u_i,p_i)\in\mathcal C}
    \mathbb 1\big[\|\bar Rp_i+\bar t-q_i\|_2<\tau_c\big].
    \label{eq:supp-ir}
\end{equation}
The reported IR is measured on the solver input: the $B_f$ pairs selected by
SC$^2$ for SparseGraph-I2P, and each baseline's own putative set, which its
PnP-RANSAC receives.

\paragraph{Feature Matching Recall (FMR).}
FMR is the fraction of pairs whose candidate set $\mathcal C_{\mathrm{cand}}$
reaches $\mathrm{IR}(\mathcal C_{\mathrm{cand}})\ge10\%$. For
SparseGraph-I2P, $\mathcal C_{\mathrm{cand}}$ is the output of Sparse Fine
Matching before fine filtering; for the baselines it is the same putative
set as for IR. FMR therefore measures whether the candidates admit a robust
estimate, not the quality of the final selection.

\paragraph{Registration Recall (RR).}
For an estimated pose $(R,t)$, the registration error is
\begin{equation}
    \mathrm{RMSE}=\Big(\frac{1}{|P|}\sum_{p\in P}
    \big\|(\bar Rp+\bar t)-(Rp+t)\big\|_2^2\Big)^{1/2},
    \label{eq:supp-rmse}
\end{equation}
and RR is the fraction of pairs with $\mathrm{RMSE}<10$\,cm. Pairs with fewer
than four correspondences count as failures.

\paragraph{Rotation and translation errors.}
For registered pairs, the relative rotation and translation errors are
\begin{equation}
    \mathrm{RRE}=\arccos\frac{\operatorname{tr}(R^\top\bar R)-1}{2},
    \qquad
    \mathrm{RTE}=\|t-\bar t\|_2 .
    \label{eq:supp-rre-rte}
\end{equation}

\paragraph{Aggregation.}
Every metric is first averaged over the pairs of a scene and then over
scenes, so each scene has equal weight. RRE and RTE are averaged over the
registered pairs of each scene only (Figure~\ref{fig:supp-pose-errors}).

\section{Datasets and Splits}
\label{sec:supp-datasets}
Both indoor benchmarks use the image--point-cloud pairs released with
2D3D-MATR~\cite{li2023matr}. Each pair links an RGB-D frame to a point-cloud
fragment of the same sequence, and pairs are kept by their image--cloud
overlap ratio. Images enter the frozen backbone in grayscale; sensor depth
is used for the depth-aware stages and for evaluation, with valid depths up
to 6\,m; coarse patches without valid depth are excluded from the candidate pool.

\paragraph{RGB-D Scenes V2.}
The dataset~\cite{lai2014unsupervised} contains 14 indoor scenes of
furniture and tabletop objects. We use scenes 1--9 for training (1857 pairs),
scene 10 for validation (127 pairs), and scenes 11--14 for testing
(497 pairs), all with overlap at least 0.3. The test pairs are identical to
those of 2D3D-MATR. The random split distributed with
Diff$^2$I2P~\cite{mu2025diff2i2p} (1978/117/386 pairs drawn from all 14
scenes) is not used: its training split contains 79.5\% of the standard test
pairs.

\paragraph{7-Scenes.}
The dataset~\cite{shotton2013scene} contains seven indoor scenes recorded
as multiple sequences. We use the sequence-level split of 2D3D-MATR with
overlap at least 0.5, giving 4048 training, 1011 validation, and 2304 test
pairs; the test pairs come from 18 sequences.

\section{Baseline Configurations}
\label{sec:supp-baselines}
For the indoor benchmarks, all baselines keep their own registration backends and are scored by the
same evaluation script as SparseGraph-I2P. SC$^2$ filtering is used only in
our method.

\paragraph{Checkpoints.}
2D3D-MATR uses its released weights. Diff$^2$I2P uses its released weights
on 7-Scenes and is retrained on RGB-D Scenes V2 scenes 1--9 with its full
configuration, because its released RGB-D weights were trained on a split
containing 79.5\% of the standard test pairs. GLASS is trained for 30 epochs.
Ex-Sim(3)-Reg follows Eq.~(5) and Table~11 of its paper, with 0.2\,m distance
tolerance, eight scale intervals over $\pm15\%$, and 16\,px reprojection
filtering applied to 2D3D-MATR correspondences; its IR and FMR are measured after
this pruning. The Hg-I2P checkpoint is selected by validation IR and reported
without HC-pruning, the stronger variant at this checkpoint: HC-pruning lowers
validation RR from 61.7\% to 61.2\% and test RR from 59.7\% to 53.1\% on
RGB-D Scenes V2 and from 82.5\% to 81.3\% on 7-Scenes.

\paragraph{Depth inputs.}
In the main benchmark, every method receives the image-side sensor depth.
2D3D-MATR, Diff$^2$I2P, GLASS, and Hg-I2P use it as a validity mask.
FreeReg receives it in place of ZoeDepth in its geometric branch, and
Ex-Sim(3)-Reg in place of DAv2 in its pruning; both improve over their
predicted-depth variants (Figure~\ref{fig:predicted-depth}a). Their registration
backends are unchanged.

\paragraph{Latency.}
Latency includes the forward pass, filtering, and pose estimation on an
otherwise idle RTX 5090 and excludes depth prediction. Evaluation-only
computation inside a forward pass, such as generating ground-truth patch
correspondences, is removed before timing.

\section{Implementation Details}
\label{sec:supp-implementation}
These details supplement the selection and execution mechanisms described
in Section~\ref{sec:method}. The main benchmark uses sensor depth; calibration and
evaluation settings for the predicted-depth extension are given in
Section~\ref{sec:supp-predicted}.

\paragraph{Graph node features.}
Each patch-pair hypothesis has 13 input features. Six describe appearance:
coarse cosine similarity, the gaps to the best image-to-point and
point-to-image similarities, the two directional ranks divided by the pool
parameter $k_c$, and a mutual-top-1 indicator. Four describe geometry:
first-order degree, second-order degree, spectral support, and local
geometric support. First-order degree sums the soft compatibility row and
divides by the pool size. Second-order degree sums the second-order support
row; spectral support is estimated by ten power iterations on that matrix.
Local support sums soft compatibility to hypotheses whose image-side 3D
centers lie within 0.3\,m. Second-order degree, spectral support, and local
support are each divided by their maximum across nodes. Three one-hot
features encode the image pyramid level. The second-order attention bias is
also normalized by its maximum entry, retaining zero when the matrix is zero.

\paragraph{Graph architecture and optimization.}
An MLP maps node features through $13\!\to\!64\!\to\!64$ dimensions, followed
by three pre-norm graph-attention layers with four heads each (16 dimensions
per head). Each layer contains residual attention and a
$64\!\to\!128\!\to\!64$ feed-forward block. The final score head applies
layer normalization and a $64\!\to\!64\!\to\!1$ MLP. The geometric bias weights $w_1,w_2$ are learned separately for every head and layer. Training uses
class-weighted binary cross-entropy, AdamW with learning rate $10^{-3}$ and
weight decay $10^{-4}$, for 15 epochs. The checkpoint with highest validation
AUC is selected. Main-table accuracy averages graph-head seeds 7351, 1, and
2; latency and the module examples use seed 7351. Selection takes the
globally highest-scoring hypotheses without a per-image-patch quota.

\paragraph{Fine decoder.}
The point-wise decoder is a $128\!\to\!256\!\to\!128$ MLP with layer
normalization and LeakyReLU between the linear layers. It is trained for
30 epochs with the original fine-matching loss and a unit-weight cosine
distillation loss against the frozen 2D3D-MATR dense decoder. The hidden width
is chosen from $\{64,256\}$ on each dataset's validation split. 7-Scenes
selects 256; on RGB-D Scenes V2 both widths reach 95.3\% validation RR,
and 256 is kept for its higher validation IR (97.6\% versus 96.7\%) and for
consistency across datasets. Dense-query
and unique-query executions evaluate this same point-wise network; the
learned decoder is not an exact replacement for 2D3D-MATR's convolutional decoder.
In Table~\ref{tab:ablation-components}, (b)$\to$(c) changes this architecture together
with coarse selection and fine execution, so it does not identify the
decoder's effect. The (d)$\to$(f) and (e)$\to$(g) comparisons instead fix the
decoder weights, coarse selection, and backend, isolating dense versus
on-demand evaluation of the same MLP. A convolutional-versus-MLP comparison with identical hypotheses
and backend would be needed to isolate the decoder replacement itself.

\paragraph{Training objectives.}
The decoder and graph head are optimized separately with the 2D3D-MATR backbone
frozen. For graph supervision, a candidate is positive if both directional
ground-truth patch-overlap ratios are nonzero. The positive-class weight
is the total number of negative candidates divided by the total number of
positive candidates in the training split, fixed throughout graph training.
For decoder supervision, the fine circle loss follows 2D3D-MATR.
Positive pairs have both 3D distance below 3.75\,cm and reprojection distance
below 8 pixels; negative pairs have either 3D distance above 10\,cm or
reprojection distance above 12 pixels. Remaining pairs are ignored.
The circle-loss margins are 0.1 and 1.4, with log scale 24.
The unit-weight distillation term is
\begin{equation}
    \mathcal{L}_{\mathrm{distill}}=
    \frac{1}{|\mathcal{T}|}\sum_{u\in\mathcal{T}}
    \left[1-\frac{f(u)^\top f_T(u)}{\|f(u)\|_2\|f_T(u)\|_2}\right],
\end{equation}
where $\mathcal{T}$ contains image pixels from up to 256 sampled
ground-truth pixel--point correspondences per training pair, and $f_T$
is the frozen dense decoder's output. Sampling uses the ground-truth
correspondences; the query union in Section~\ref{sec:sparse} describes inference.

\paragraph{Tree indices and GPU scheduling.}
Image blocks form a three-level quadtree stored as level and grid-index
arrays. The parent of $(\ell,a,b)$ is
$(\ell+1,\lfloor a/2\rfloor,\lfloor b/2\rfloor)$. All levels use the same
stride-2 pixel lattice, so a parent contains its children's sampled pixels.
With $N_s$ point superpoints, pair keys $rN_s+s$ support two vectorized
ancestor passes. A child pair is removed only if an ancestor is paired with
the same superpoint. The unique-pixel descriptor bank is constructed from
the selected set before pruning. Remaining blocks are grouped into at most
three level batches and matched by padded, masked batched matrix products.
Pixel--point IDs are deduplicated after patch-wise mutual top-$k_f$
selection. Pruning preserves the union of compared edges, but can change
local mutual-match rankings; it is not a guarantee of identical final
correspondences or registration recall.

\paragraph{GPU pseudocode.}
Algorithm~\ref{alg:sparse-fine} uses the notation of Section~\ref{sec:method}:
$\mathcal{H}$ is the selected patch-pair set, $U_r$ is a patch's sampled
pixel set, and $\mathcal{P}_s$ is a point block. $\operatorname{Decode}$
applies bilinear sampling and the point-wise MLP. The lookup table $\Lambda$
indexes the shared descriptor bank $F_{\mathcal U}=[f(u)]_{u\in\mathcal U}$. $X_\ell,Y_\ell$ contain gathered pixel
and point descriptors from $F_P^f$; $V_\ell$ masks padding. Grouping blocks by level
keeps their pixel dimensions uniform for batched products.

\begin{algorithm}[t]
\caption{Sparse Fine Matching on GPU}
\label{alg:sparse-fine}
\footnotesize
\begin{algorithmic}[1]
\Require $\mathcal{H}$, tree indices, $\{U_r,\mathcal{P}_s\}$, $F^{1/2}$, point descriptors $F_P^f$
\Ensure Fine pixel--point correspondences $\mathcal{C}$
\State $\mathcal{U}\gets\operatorname{Unique}(\operatorname{Concat}_{(r,s)\in\mathcal{H}}U_r)$
\State $F_{\mathcal{U}}\gets\operatorname{Decode}(F^{1/2},\mathcal{U})$; build $\Lambda$
\State $\mathcal{K}\gets\{rN_s+s:(r,s)\in\mathcal{H}\}$; $\mu\gets\mathbf{true}$
\For{$a=1,2$} \Comment{two vectorized ancestor passes}
    \State $\mu\gets\mu\land\neg\operatorname{CoveredByAncestor}(\mathcal{H},a,\mathcal{K})$
\EndFor
\State $\{\mathcal{H}^{\star}_{\ell}\}\gets\operatorname{GroupByLevel}(\mathcal{H}[\mu])$; $\mathcal{C}\gets\emptyset$
\For{each nonempty level $\ell$}
    \State $(X_{\ell},Y_{\ell},V_{\ell})\gets\operatorname{Gather}(F_{\mathcal{U}},\Lambda,F_P^f,\mathcal{H}^{\star}_{\ell})$
    \State $Z_{\ell}\gets\tfrac12(\mathbf{1}+\operatorname{BatchedMatMul}(X_{\ell},Y_{\ell}^{\top}))$
    \State $E_{\ell}\gets\operatorname{MutualTopK}(Z_{\ell},V_{\ell},k_f,\gamma)$
    \State $\mathcal{C}\gets\mathcal{C}\cup\operatorname{GlobalPixelPointIDs}(E_{\ell})$
\EndFor
\State \Return $\operatorname{Unique}(\mathcal{C})$
\end{algorithmic}
\end{algorithm}

\paragraph{Sensor-depth spectral selection and pose objective.}
For the fine-correspondence set $\mathcal C=\{(u_i,p_i)\}$, lift pixels to
$q_i=D(u_i)K^{-1}\widetilde u_i$. Let $C$ be their soft distance
compatibility matrix, $A_{ij}=\mathbb{1}[C_{ij}>\tau]$, and
$S=(AA)\odot A$.
From $v^{(0)}=\mathbf{1}/\sqrt{|\mathcal C|}$, power iteration estimates spectral support:
\begin{equation}
    v^{(n+1)}=\frac{S v^{(n)}}{\|S v^{(n)}\|_2},
    \qquad n=0,\ldots,9.
    \label{supp-eq:fine-spectral}
\end{equation}
Using uniform scores if normalization degenerates, we select
\begin{equation}
    \mathcal C^{\star}=\{(u_i,p_i):i\in\operatorname{TopK}_{B_f}(|v^{(10)}|)\}.
    \label{supp-eq:fine-selection}
\end{equation}
Here $B_f$ is the solver-input budget; TopK returns at most the available
number of pairs. The selected 2D--3D pairs define the PnP reprojection objective
\begin{equation}
    E(R,t)=\sum_{(u_i,p_i)\in\mathcal C^{\star}}
    \left\|\pi\!\left(K(Rp_i+t)\right)-u_i\right\|_2^2,
    \label{eq:epnp-reprojection}
\end{equation}
where $\pi([x,y,z]^\top)=(x/z,y/z)^\top$.

EPnP estimates the pose from the selected original pixel--point pairs and
intrinsics, without using the lifted coordinates as solver inputs.

\section{Additional Experiments}
\label{sec:supp-experiments}

\subsection{Rotation and Translation Errors}
\label{sec:supp-pose-errors}
\begin{figure*}[t]
    \centering
    \includegraphics[width=\textwidth]{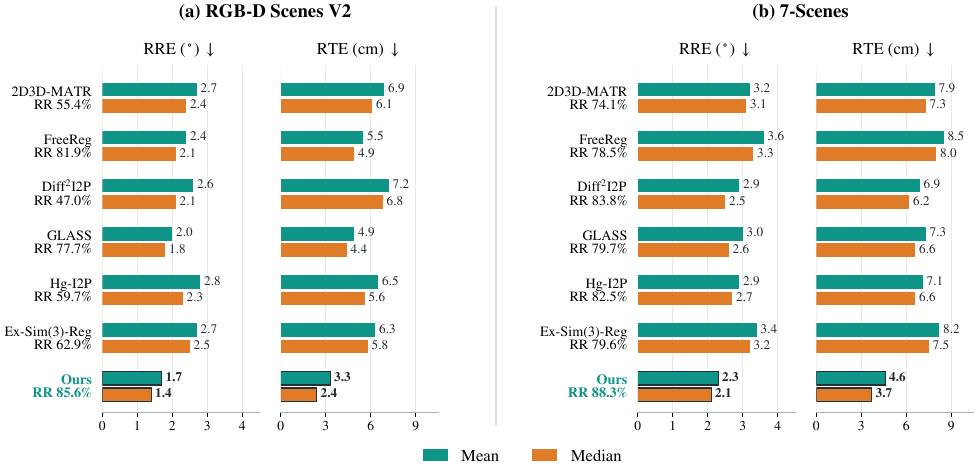}
    \caption{Rotation and translation errors on (a) RGB-D Scenes V2 and
    (b) 7-Scenes. Teal/orange bars show mean/median RRE ($^\circ$) and RTE
    (cm) over registered pairs (RMSE $<10$\,cm). Each statistic is computed
    within a scene and averaged equally over scenes; ours further averages
    three graph-head seeds. Row labels give RR (\%), the fraction of pairs
    covered by each method's error statistics. These statistics are
    conditional on each method's registered subset.}
    \label{fig:supp-pose-errors}
\end{figure*}

Figure~\ref{fig:supp-pose-errors} complements the RR of Tables~\ref{tab:rgbd-evaluation}
and~\ref{tab:7scenes-evaluation} with RRE and RTE (Eq.~\eqref{eq:supp-rre-rte}). SparseGraph-I2P has
the lowest mean and median errors on both datasets. Its median RTE is
2.4\,cm on RGB-D Scenes V2, against 4.4\,cm for GLASS, and 3.7\,cm on
7-Scenes, against 6.2\,cm for Diff$^2$I2P, the next lowest values. Because
our averages cover the largest share of pairs, the lower errors do not come
from averaging over fewer registered pairs.

\subsection{Outdoor Registration on KITTI-DC}
\label{sec:supp-kitti-dc}

\paragraph{Data and depth inputs.}
We evaluate outdoor image-to-point-cloud registration on the KITTI-DC
cross-frame benchmark used by FreeReg~\cite{wang2024freereg} and
Diff$^2$I2P~\cite{mu2025diff2i2p}. The test set contains 342 pairs from
four drives (2011\_09\_26: 0005, 0020, 0079, and 0113). Each query image
is paired with a LiDAR scan from one of the next three frames, within
10\,m. The training set contains 3,855 pairs from 138 disjoint drives;
validation uses 272 pairs from three further drives. Learned methods use
a $224\times1216$ image crop and a 0.25\,m point-cloud voxel size, with
depths in $[0.5,80]$\,m. Image-side input depth comes from the query
frame's own LiDAR scan, completed with IP-Basic. The target point cloud
comes from a different scan. All compared methods receive the same
image-side depth, including FreeReg in place of its ZoeDepth predictions.
FreeReg runs at its native $1242\times225$ resolution, with output pixels
mapped to the common crop for evaluation.
Ground-truth poses are computed from OXTS poses and KITTI calibration
extrinsics, rather than FreeReg's approximate axis conversion; the numbers
therefore should not be compared directly with published results that use
the latter ground truth.

\paragraph{Outdoor evaluation protocol.}
The indoor thresholds of Section~\ref{sec:supp-metrics} are replaced by
the KITTI-DC thresholds. A correspondence is an inlier when its 3D
residual is below 3\,m, using the benchmark's dense evaluation depth to
lift image pixels; correspondences without evaluation depth are excluded
from IR. FMR uses a 5\% inlier-ratio threshold on the unfiltered candidate
set, whereas IR is measured on the solver input. A pair is registered
when both $\mathrm{RRE}<10^\circ$ and $\mathrm{RTE}<3$\,m. Metrics are
averaged within each test drive and then across the four drives.
Table~\ref{tab:kitti-dc} reports mean RRE and RTE both over all test pairs
and over registered pairs. The latter averages cover different subsets
for different methods and should be read together with RR.

\paragraph{Outdoor configuration.}
The learned methods are trained on the KITTI-DC training split.
SparseGraph-I2P freezes the KITTI-trained 2D3D-MATR backbone and learns its
sparse fine decoder and Coarse Hypothesis Graph. The image patch grid is
$16\times64$ at the finest level; the coarse budget remains 1,024 patch
pairs. The graph compatibility scale and local-support radius are
1.0\,m and 3.0\,m, respectively. Fine SC$^2$ parameters are selected on
validation from $\sigma_f\in\{0.0625,0.125,0.25,0.5\}$\,m and
$B_f\in\{64,128,256\}$. Baselines use PnP-RANSAC with 50,000 iterations;
our method uses fine SC$^2$ selection followed by a single EPnP call.

\begin{table*}[t]
\centering\footnotesize
\caption{Outdoor registration on KITTI-DC (342 test pairs, four drives).
FMR, IR, and RR are in percent; RRE is in degrees and RTE in meters.
All includes every test pair; Registered includes only
pairs with RRE $<10^\circ$ and RTE $<3$\,m.
Best values are bold, including ties.}
\label{tab:kitti-dc}
\setlength{\tabcolsep}{5pt}
\renewcommand{\arraystretch}{1.05}
\begin{tabular*}{\textwidth}{@{\extracolsep{\fill}}lrrrrrrr@{}}
\toprule
 & & & & \multicolumn{2}{c}{All} & \multicolumn{2}{c}{Registered} \\
\cmidrule(lr){5-6}\cmidrule(lr){7-8}
Method & FMR$\uparrow$ & IR$\uparrow$ & RR$\uparrow$ & RRE$\downarrow$ & RTE$\downarrow$ & RRE$\downarrow$ & RTE$\downarrow$ \\
\midrule
FreeReg & \textbf{100.0} & 60.4 & 65.8 & 8.21 & 2.91 & 4.64 & 1.62 \\
Diff$^2$I2P & 99.7 & 66.7 & 77.5 & \textbf{2.79} & \textbf{1.84} & \textbf{1.88} & \textbf{1.10} \\
2D3D-MATR & 98.8 & \textbf{89.7} & 71.6 & 9.65 & 3.13 & 3.06 & 1.27 \\
\midrule
\textbf{Ours} & 99.4 & 81.0 & \textbf{85.1} & 3.71 & \textbf{1.84} & 2.53 & 1.36 \\
\bottomrule
\end{tabular*}
\end{table*}

\paragraph{Results.}
SparseGraph-I2P reaches 85.1\% RR, improving on Diff$^2$I2P by 7.6
percentage points, 2D3D-MATR by 13.5 points, and FreeReg by 19.3 points.
Its solver-input IR is 81.0\%, below 2D3D-MATR's 89.7\%, despite the
higher registration recall. Thus correspondence purity alone does not
determine pose success. All methods have FMR of at least 98.8\%, so the
recall gap arises despite most pairs passing the candidate-level inlier
test. Diff$^2$I2P has the lowest mean RRE over all pairs and the lowest
RRE and RTE over its registered subset; it ties our all-pair RTE at
1.84\,m. These results support higher outdoor registration recall for
our method under this depth-assisted protocol, without a uniform advantage
in correspondence purity or conditional pose accuracy.

\section{Registration with Predicted Depth}
\label{sec:supp-predicted}

\subsection{Calibration and Filtering}
\label{sec:supp-depth-calibration}
This section derives the predicted-depth path summarized in
Section~\ref{sec:backend}. Inputs are the fine candidates $\{(u_i,p_i)\}_{i=1}^{N}$
produced by Sparse Fine Matching, the predicted depth $d_i=D(u_i)$, and the
intrinsics $K$. Neither sensor depth nor the pose is used.

\paragraph{Why sensor-depth SC$^2$ fails.}
Let $r_i=K^{-1}\widetilde u_i$ be the unit-depth ray of $u_i$, so that
$q_i=z_ir_i$ for true depth $z_i$. Suppose the prediction has a global scale
error, $d_i=z_i/s$. For a correct pair $(i,j)$, the lifted length is
$\|d_ir_i-d_jr_j\|_2=\|p_i-p_j\|_2/s$, and the distance residual in
Eq.~\eqref{eq:backprojection} becomes
\begin{equation}
    \big|\|d_ir_i-d_jr_j\|_2-\|p_i-p_j\|_2\big|
    =\big|1-s^{-1}\big|\,\|p_i-p_j\|_2 .
    \label{eq:supp-scale-residual}
\end{equation}
The residual grows linearly with the pair length. With the sensor-depth
tolerance $\sigma_f=6.25$\,mm on RGB-D Scenes V2, a $1\%$ scale error
already exceeds $\sigma_f$ for correct pairs longer than $0.63$\,m, so true
correspondences lose their compatible neighbors and the spectral ranking
no longer favors them. An offset in depth or inverse depth adds a
depth-dependent term to the same residual. The path below therefore
removes the scale before coarse selection, calibrates an affine model
before fine filtering, and widens the tolerance in proportion to length.

\begin{figure*}[t]
    \centering
    \includegraphics[width=\textwidth]{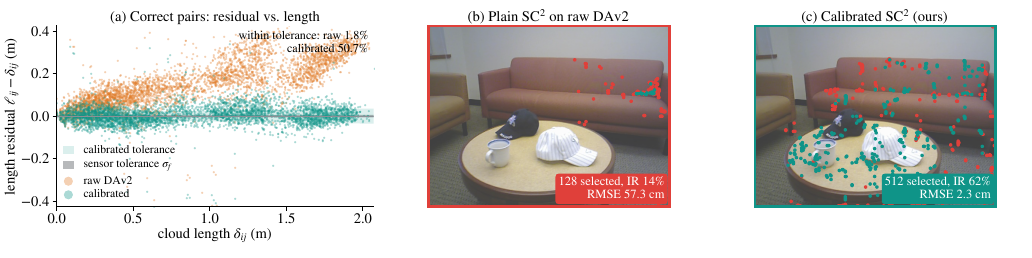}
    \caption{SC$^2$ filtering with DAv2 depth on an RGB-D Scenes V2 test pair.
    (a)~Length residuals of ground-truth-correct candidate pairs grow with
    length under raw depth, as in Eq.~\eqref{eq:supp-scale-residual}, and
    center on zero after calibration. (b)~With the sensor-depth settings,
    SC$^2$ keeps a compact cluster and registration fails. (c)~The calibrated
    path selects spatially spread correspondences and registers the pair.
    Green: inlier within 5\,cm; red: outlier. Ground truth is used for
    coloring only.}
    \label{fig:supp-sc2-calibration}
\end{figure*}

\begin{figure*}[t]
    \centering
    \includegraphics[width=\textwidth]{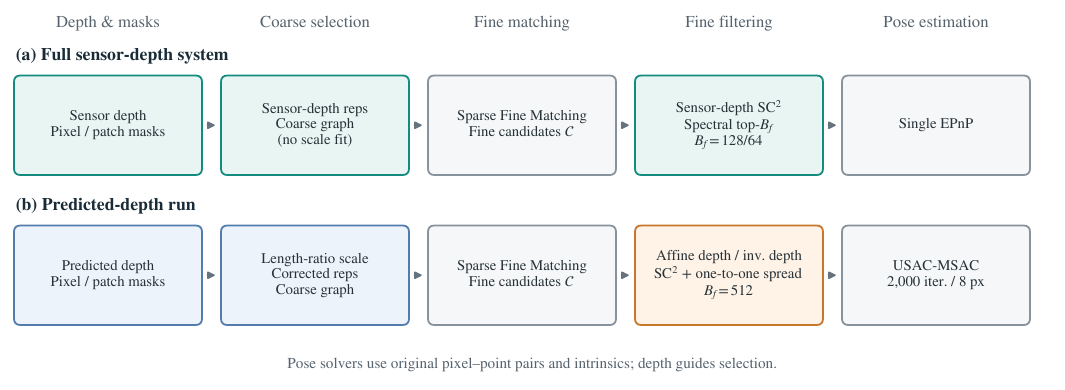}
    \caption{Depth sources and execution order in SparseGraph-I2P.
    (a) Full sensor-depth system. (b) Predicted-depth inference, with no
    image-side sensor depth: scale correction precedes coarse graph
    selection, and affine depth/inverse-depth calibration precedes fine
    filtering. No affine fine-depth calibration is applied in (a).
    Budgets 128/64 refer to RGB-D Scenes V2/7-Scenes.
    All configurations use RGB, the point cloud, and intrinsics;
    ground-truth poses are used only for scoring.}
    \label{fig:supp-depth-provenance}
\end{figure*}

Figure~\ref{fig:supp-sc2-calibration} illustrates this effect. On the
RGB-D Scenes V2 test set with DAv2 depth and solver seed 7351, SC$^2$ with
the sensor-depth settings registers 2.9\% of pairs (scene mean), whereas
the calibrated path below registers 78.7\%.

\paragraph{Coarse-stage scale.}
For coarse hypotheses with representatives $(q_i,p_i)$ lifted by $D$, a
single scale is estimated from length ratios
$\rho_{ij}=\|p_i-p_j\|_2/\|q_i-q_j\|_2$ over $32{,}768$ random pairs with
both lengths above $0.15$\,m and $\rho_{ij}\in[0.25,4]$. Under
Eq.~\eqref{eq:supp-scale-residual}, every correct pair yields
$\rho_{ij}=s$, whereas incorrect pairs spread over the range. The estimate
\begin{equation}
    \hat s=\exp\!\Big(\operatorname*{median}_{(i,j):\,|\log\rho_{ij}-c|<0.06}
    \log\rho_{ij}\Big)
    \label{eq:supp-coarse-scale}
\end{equation}
refines the center $c$ of the highest bin of a $160$-bin histogram of
$\log\rho_{ij}$, smoothed with the kernel $[1,2,3,2,1]$. The Coarse Hypothesis
Graph then uses $\hat s\,q_i$ with $\sigma_c=0.15$\,m; its weights are
unchanged from sensor-depth training.

\paragraph{Affine depth models.}
Let $m$ be the median of $\{d_i\}$. Two families map $d_i$ to calibrated
depth, affine in inverse depth (inv) and in depth (aff):
\begin{equation}
    \begin{aligned}
    \hat z_i^{\mathrm{inv}}(\alpha,\beta)&=
    \frac{\alpha\,m}{(1-\beta)\,m/d_i+\beta},\\
    \hat z_i^{\mathrm{aff}}(\alpha,\beta)&=
    \alpha\big[(1-\beta)\,d_i+\beta\,m\big].
    \end{aligned}
    \label{eq:depth-calibration}
\end{equation}
Here $\alpha$ scales the median depth and $\beta$ sets the offset;
$\beta=0$ reduces both families to $\hat z_i=\alpha d_i$. Candidates with
non-positive denominators or $\hat z_i<0.05$\,m are invalid. Both families
are evaluated for every pair; neither is assigned to a predictor.

\paragraph{Length factorization and voting.}
With $\hat q_i=\hat z_i r_i$ and
$\ell_{ij}(\alpha,\beta)=\|\hat q_i-\hat q_j\|_2$,
Eq.~\eqref{eq:depth-calibration} is linear in $\alpha$, hence
\begin{equation}
    \ell_{ij}(\alpha,\beta)=\alpha\,\ell_{ij}(1,\beta).
    \label{eq:supp-length-factorization}
\end{equation}
For a fixed $\beta$, every $\alpha$ is therefore scored from one set of unit
lengths. Each family is initialized by maximizing a soft length vote over a
set $\Omega$ of $40{,}000$ random candidate pairs with
$\delta_{ij}=\|p_i-p_j\|_2>0.15$\,m:
\begin{equation}
    (\alpha^0,\beta^0)=\arg\max_{\alpha,\beta}\sum_{(i,j)\in\Omega}
    \Big[1-\Big(\frac{\alpha\,\ell_{ij}(1,\beta)-\delta_{ij}}
    {\tau_{ij}}\Big)^{2}\Big]_+,
    \label{eq:supp-vote}
\end{equation}
with $\tau_{ij}=0.02\,\mathrm{m}+0.02\,\delta_{ij}$, $\alpha$ on $97$
log-spaced values in $[0.4,2.5]$, and $\beta$ on $37$ values in
$[-0.9,0.9]$. The vote needs neither a pose nor sensor depth.

\paragraph{Model selection by the leading eigenvalue.}
The vote counts pairs independently, so outlier pairs can still bias it.
Each hypothesis $h=(f,\alpha,\beta)$ is therefore rescored by the
collective consistency it induces. With
\begin{equation}
    C_{ij}(h)=\Big[1-\Big(\frac{\ell_{ij}(h)-\delta_{ij}}{\sigma_{ij}}\Big)^{2}\Big]_+,
    \quad \sigma_{ij}=0.015\,\mathrm{m}+0.01\,\delta_{ij},
    \label{eq:supp-calibrated-compatibility}
\end{equation}
$C_{ii}=0$, and $C_{ij}=0$ for $\delta_{ij}\le0.01$\,m, the objective is
\begin{equation}
    \lambda(h)=\lambda_{\max}\big((C(h)C(h))\odot C(h)\big).
    \label{eq:supp-eigen-objective}
\end{equation}
For a cluster of $n$ candidates that are mutually compatible with
$C_{ij}=1$ and isolated from the rest, every off-diagonal entry of $CC$
within the cluster equals $n-2$. The cluster block of the second-order
matrix is then $(n-2)(\mathbf 1\mathbf 1^\top-I)$, with leading eigenvalue
$(n-1)(n-2)$. Thus $\lambda(h)$ grows quadratically with the size of the
largest consistent cluster, and the soft $C$ makes it peak at exact length
agreement. Incorrect candidates outside the cluster barely affect
$\lambda(h)$, unlike a least-squares fit over all candidates. The search
evaluates $\lambda$ on $H_1=\{(f,\alpha_f^0e^{t},\beta_f^0+b)\}$ with
$f\in\{\mathrm{inv},\mathrm{aff}\}$, $13$ values of $t\in[-0.12,0.12]$,
and $5$ values of $b\in[-0.3,0.3]$, then on a finer grid around the best
hypothesis of the same family ($5$ values each of $t\in[-0.02,0.02]$ and
$b\in[-0.06,0.06]$). The overall maximizer is $h^\star$. Each $\lambda$ uses
$15$ power iterations on at most $1{,}500$ randomly subsampled candidates,
and $\beta$ is clipped to $[-1.5,0.95]$.

\paragraph{Fine filtering.}
The calibrated points $\hat q_i=\hat z_i(h^\star)r_i$ replace $q_i$. By
Eq.~\eqref{eq:supp-scale-residual}, a residual scale error $\epsilon$ leaves
a residual proportional to length, so the tolerance gains a relative term:
\begin{equation}
    \sigma_{ij}=\sigma_{\mathrm{abs}}+\sigma_{\mathrm{rel}}
    \max\!\big(\|\hat q_i-\hat q_j\|_2,\delta_{ij}\big),
    \label{eq:predicted-tolerance}
\end{equation}
with $\sigma_{\mathrm{abs}}=0.015$\,m and $\sigma_{\mathrm{rel}}=0.01$, which
admits $|\epsilon|$ up to $1\%$ on long pairs. Compatibility then follows
Eq.~\eqref{eq:backprojection}, and
$A_{ij}=\mathbb 1[C_{ij}>\tau]\,\mathbb 1[\|\hat q_i-\hat q_j\|_2>0.01]\,
\mathbb 1[\delta_{ij}>0.01]$, where the length conditions exclude pairs
that share a pixel or a point. The leading eigenvector $v$ of
$S=(AA)\odot A$ (10 power iterations) ranks the candidates.

\paragraph{Spatially spread selection.}
The ray coordinates $(r_{i,x},r_{i,y})$ are normalized to $[0,1]^2$ over
the candidates and binned on a $G\times G$ grid with $G=8$. Let $O$ be the
number of occupied cells. Candidates are visited in decreasing $v_i$ and
accepted if neither their pixel nor their point has been used and their
cell holds fewer than
\begin{equation}
    Q=\big\lceil B_f/O\big\rceil
    \label{eq:supp-quota}
\end{equation}
accepted candidates, with $B_f=512$. A second pass without the quota fills
the remaining budget. The quota keeps the leading spectral cluster from
filling the whole budget with nearby candidates, which constrain PnP poorly.

\paragraph{Pose estimation.}
USAC-MSAC estimates the pose from the original 2D
pixels and 3D points of the selected set, with $2{,}000$ iterations and an
8\,px reprojection threshold. Calibrated depth is used only for selection.

\subsection{Evaluation Details}
\label{sec:supp-predicted-depth}
The predictions used in Figure~\ref{fig:predicted-depth} come
from DAv2 (Hypersim, ViT-L) and DA3 (Metric-Large). For the predicted-depth
runs, inference receives only RGB, intrinsics, and the point cloud;
sensor depth and ground-truth poses are used only for scoring.
Figure~\ref{fig:supp-depth-provenance} shows the depth sources and execution order
of our method. No model is retrained. Our method uses the
sensor-trained backbone and Coarse Hypothesis Graph (seed 7351), with the
calibrated filter and robust solver described in
Section~\ref{sec:backend}. Baselines retain their main-table checkpoints and solvers and receive
the same depth predictions wherever they read image depth; FreeReg replaces
its ZoeDepth branch. Baselines are also rerun with sensor depth in the same
session, reproducing main-table recall within 1.0 point.

\paragraph{Sensor-fine diagnostic reference.}
Our dashed reference deliberately adds sensor depth for fine filtering.
It retains the predicted-depth coarse hypotheses and pre-filter fine
candidates, bypasses affine calibration, and applies the sensor-depth
SC$^2$ settings: $\sigma_f=6.25/12.5$\,mm and $B_f=128/64$ on
RGB-D Scenes V2/7-Scenes, without the spatial quota. It retains the
USAC-MSAC solver (2,000 iterations, 8\,px) used in the predicted-depth
study. Thus it diagnoses fine-stage depth sensitivity; it is neither a
fully predicted-depth run nor the full sensor-depth system of the main
tables. Its RGB-D RR is 84.93/84.49\% for the DAv2/DA3-based candidates,
whereas the full sensor-depth system reaches 85.6\%.

\paragraph{Calibration settings.}
All settings in Section~\ref{sec:supp-depth-calibration} are fixed on
RGB-D validation (scene 10), without 7-Scenes tuning.

\paragraph{Additional accuracy observations.}
On the full 7-Scenes split, our method reaches 81.7\% RR with both
predictors, ahead of GLASS and Ex-Sim(3)-Reg (80.3\%) and 1.1 points below
Hg-I2P (82.8\%). Its drops from the sensor-fine diagnostic references are
5.8 points on RGB-D and 6.1 points on 7-Scenes with DAv2. Without calibrated
filtering, DAv2 recall drops to 69.8\% and 79.9\%, respectively. 2D3D-MATR,
GLASS, and Hg-I2P use depth only as a validity mask; dense predictions give
identical DAv2/DA3 matches and recall stays within 0.6 points of sensor depth.
FreeReg, Diff$^2$I2P, and Ex-Sim(3)-Reg use depth values and lose up to
41, 12, and 12 points. These predicted-depth runs use one graph-head seed
and a robust solver, unlike the main tables' three-seed EPnP results.

\paragraph{Runtime protocol.}
The predicted-depth figure sums separately measured depth prediction, forward, filtering,
and pose-estimation stages, counting the shared predictor once per pair.
These are not directly timed end-to-end measurements. Deployable forwards
are measured on an idle RTX 5090 using 15 test pairs per scene after ten
warm-up pairs, with evaluation-only node-label generation removed and
correspondence outputs checked. Predictor and backend costs use existing
stage records; loading, data transfers, disk IO, and offline MobileSAM are
excluded. Calibration alone costs about 43--69\,ms per pair in these
scene-mean records. Our stage sums on 7-Scenes are 163.2/105.5\,ms with
DAv2/DA3. Calibration increases cost relative to the sensor-depth system;
we do not claim the lowest runtime under predicted depth.

\section{Additional Qualitative Results}
\label{sec:supp-gallery}

Figures~\ref{fig:supp-gallery-rgbd}--\ref{fig:supp-gallery-7s-b} extend
Figure~\ref{fig:qualitative} with the same cell layout and the same
sensor-depth settings. For each test scene, one pair is drawn uniformly at
random (seed 7351) from the pairs that every method evaluates, excluding the
pairs in Figures~1 and~7, that our method registers, and whose solver input
contains at least one outlier. The last condition skips trivial cases: the
top-128/64 SC$^2$ selection consists entirely of inliers for 17--48\% of
the pairs per RGB-D Scenes V2 scene and 19--99\% per 7-Scenes scene
(graph-head seed 7351). Baseline results are not consulted; per-scene
recall of all methods is reported in Tables~\ref{tab:rgbd-evaluation} and~\ref{tab:7scenes-evaluation}.

On these pairs, our solver input contains 88--99\% inliers. The baselines'
solver inputs contain 3--97\% inliers, and their PnP-RANSAC solvers register
51 of the 66 baseline cells.

\newcommand{\gallerycaption}{Green: inlier within 5\,cm under the ground
truth; red: outlier; up to 250 lines per baseline, all 128/64 for ours.
Badges give solver-input IR; the box gives RRE\,/\,RTE (green frame: RMSE
$<10$\,cm). Row labels give the scene and the image/cloud frame indices.}

\begin{figure*}[p]
    \centering
    \includegraphics[width=\textwidth]{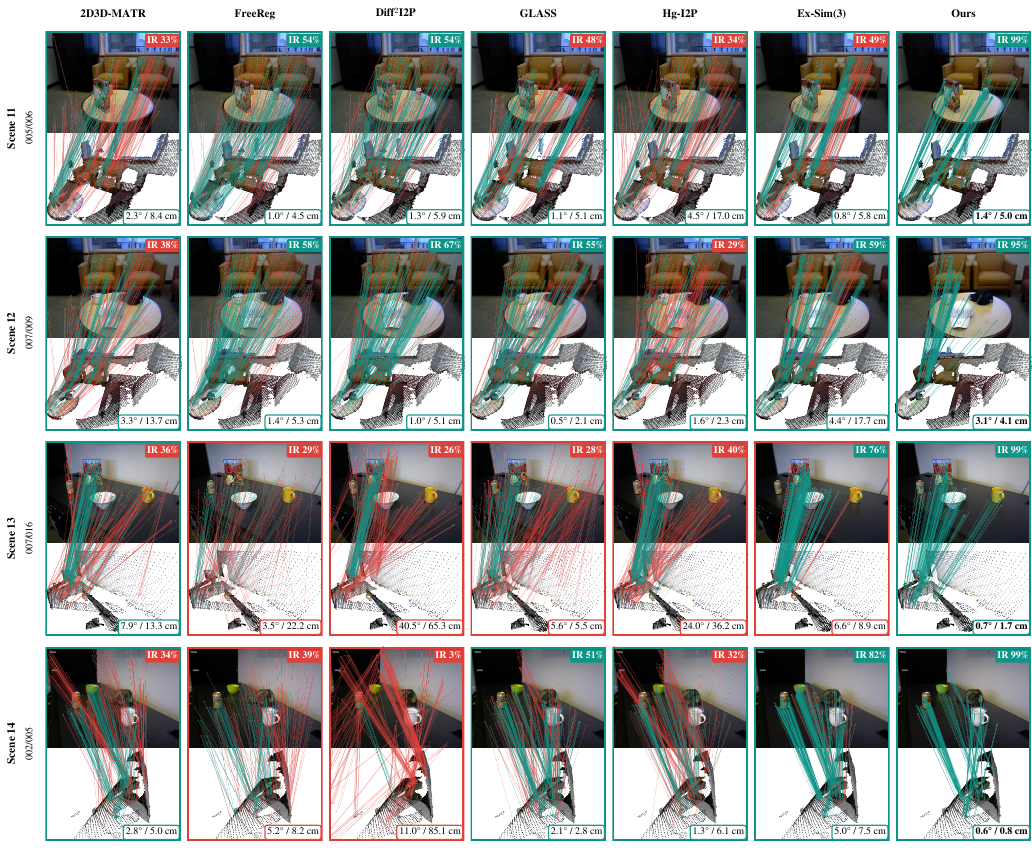}
    \caption{RGB-D Scenes V2: one random test pair per scene.
    \gallerycaption}
    \label{fig:supp-gallery-rgbd}
\end{figure*}

\begin{figure*}[p]
    \centering
    \includegraphics[width=\textwidth]{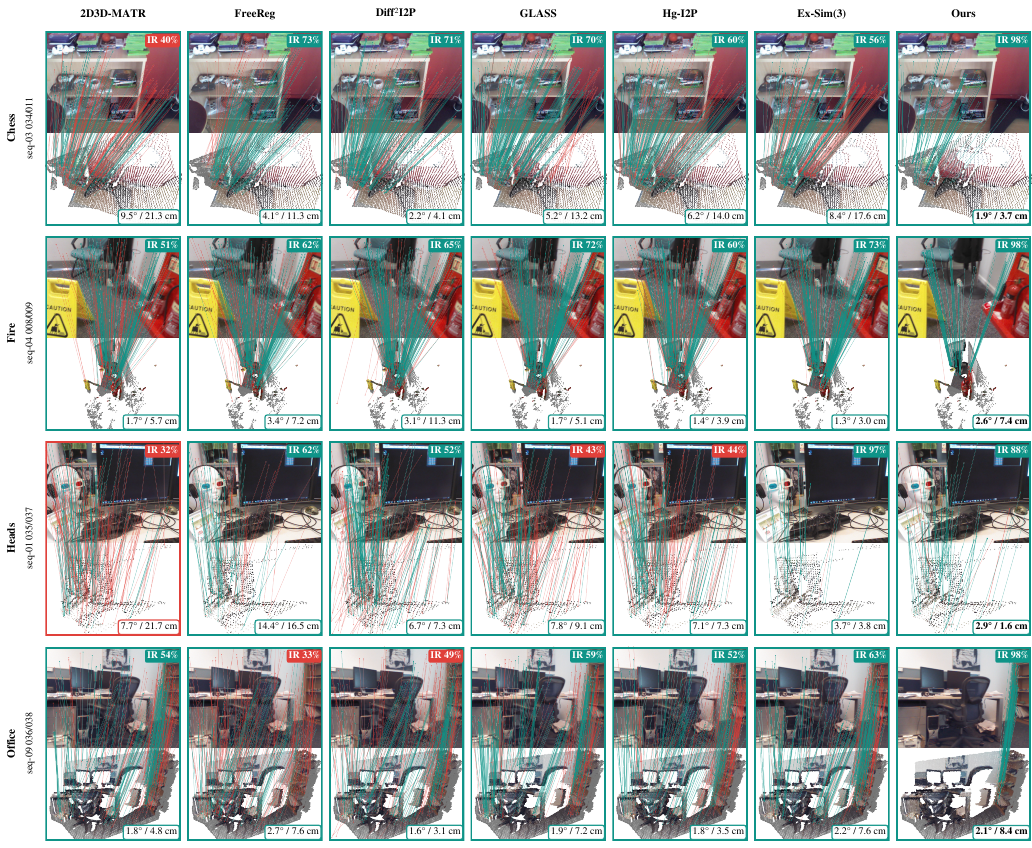}
    \caption{7-Scenes: one random test pair per scene (Chess, Fire, Heads,
    Office). \gallerycaption}
    \label{fig:supp-gallery-7s-a}
\end{figure*}

\begin{figure*}[p]
    \centering
    \includegraphics[width=\textwidth]{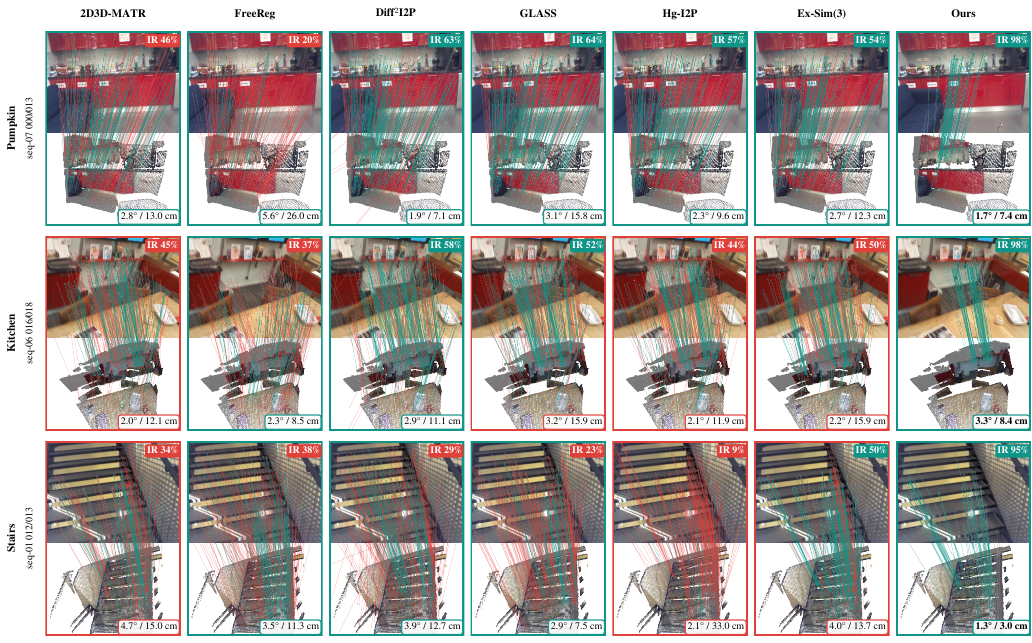}
    \caption{7-Scenes: one random test pair per scene (Pumpkin, Kitchen,
    Stairs). \gallerycaption}
    \label{fig:supp-gallery-7s-b}
\end{figure*}

\end{document}